%% file: sample-authordraft.tex
\documentclass[sigconf]{acmart}

\renewcommand\footnotetextcopyrightpermission[1]{} 
\usepackage{algorithm}
\usepackage{microtype}
\usepackage{algcompatible,amsmath}
\usepackage{graphicx}
\usepackage{url}
\usepackage{multirow}
\usepackage{comment}
\usepackage{pifont}
\usepackage{balance}
\usepackage{mathtools}
\usepackage{xcolor}
\usepackage{colortbl}
\usepackage{bbm}
\usepackage{enumitem}
\usepackage{marvosym}
\usepackage{textcomp}
\usepackage{array}
\usepackage{booktabs}
\usepackage{tabularx}
\usepackage{CJKutf8}
\usepackage{algorithmicx}  
\usepackage{algpseudocode} 

\usepackage{fancyhdr}
\usepackage{newfloat}
\usepackage{listings}
\floatstyle{ruled}
\newfloat{listing}{tb}{lst}{}
\floatname{listing}{Listing}

\newcommand{\squishlist}{
\begin{list}{$\bullet$}
{ \usecounter{Lcount}
\setlength{\itemsep}{0pt}
\setlength{\parsep}{0pt}
\setlength{\topsep}{0pt}
\setlength{\partopsep}{0pt}
\setlength{\leftmargin}{2em}
\setlength{\labelwidth}{1.5em}
\setlength{\labelsep}{0.5em} } }
\newcommand{\squishend}{
\end{list} }
\newcommand*\circled[1]{\kern-2.5em%
  \put(0,4){\color{white}\circle*{18}}\put(0,4){\circle{10}}%
  \put(-3,0){\color{black}\bfseries#1}~~}
\usepackage{tikz} 

\usepackage{enumitem}
\makeatletter
\newcommand{\thickhline}{%
    \noalign {\ifnum 0=`}\fi \hrule height 2pt
    \futurelet \reserved@a \@xhline
}

\definecolor{codegreen}{rgb}{0,0.6,0}
\definecolor{codegray}{rgb}{0.5,0.5,0.5}
\definecolor{codepurple}{rgb}{0.58,0,0.82}
\definecolor{backcolour}{rgb}{0.95,0.95,0.92}

\lstdefinestyle{mystyle}{
    backgroundcolor=\color{backcolour},   
    commentstyle=\color{codegreen},
    keywordstyle=\color{magenta},
    numberstyle=\tiny\color{codegray},
    stringstyle=\color{codepurple},
    basicstyle=\ttfamily\footnotesize,
    breakatwhitespace=false,         
    breaklines=true,                 
    captionpos=b,                    
    keepspaces=true,                 
    numbers=left,                    
    numbersep=5pt,                  
    showspaces=false,                
    showstringspaces=false,
    showtabs=false,                  
    tabsize=2
}

\copyrightyear{}
\acmYear{}
\acmDOI{}

\acmConference[SIGIR '2024]{47th International ACM SIGIR Conference on Research and Development in Information Retrieval}{July 14-18,
  2024}{Washington D.C., USA}
\acmPrice{}
\acmISBN{}

\begin{document}

\title[Give Me More Details: Improving Fact-Checking with Latent Retrieval]{Give Me More Details: Improving Fact-Checking \\ with Latent Retrieval}



\author{Xuming Hu}
\affiliation{%
  \institution{HKUST(GZ)}
  \city{Guangzhou}
  \country{China}
}
\email{xuminghu97@gmail.com}

\author{Junzhe Chen}
\affiliation{%
  \institution{Tsinghua University}
  \city{Beijing}
  \country{China}
}
\email{chenjz20@mails.tsinghua.edu.cn}

\author{Zhijiang Guo}
\authornote{Corresponding author.}
\affiliation{%
  \institution{University of Cambridge}
  \city{Cambridge}
  \country{UK}
  }
\email{zg283@cam.ac.uk}

\author{Philip S. Yu}
\affiliation{%
 \institution{University of Illinois at Chicago}
 \city{Chicago}
 \country{USA}
 }
\email{psyu@cs.uic.edu}


\begin{abstract}
Evidence plays a crucial role in automated fact-checking. When verifying real-world claims, existing fact-checking systems either assume the evidence sentences are given or use the search snippets returned by the search engine. Such methods ignore the challenges of collecting evidence and may not provide sufficient information to verify real-world claims. Aiming at building a better fact-checking system, we propose to incorporate full text from source documents as evidence and introduce two enriched datasets. The first one is a multilingual dataset, while the second one is monolingual (English). We further develop a latent variable model to jointly extract evidence sentences from documents and perform claim verification. Experiments indicate that including source documents can provide sufficient contextual clues even when gold evidence sentences are not annotated. The proposed system is able to achieve significant improvements upon best-reported models under different settings. 
\end{abstract}




\maketitle
\input{subfiles/1_intro}
\input{subfiles/2_related}

\input{subfiles/3_dataset}

\input{subfiles/4_model}

\input{subfiles/5_experiment}

\input{subfiles/8_limitation}
\input{subfiles/6_conclusion}
\balance

\bibliographystyle{ACM-Reference-Format}
\bibliography{sample-base}










\end{document}

%% file: subfiles/1_intro.tex
\section{Introduction}
\label{intro}

\begin{figure}
    \centering
    \includegraphics[scale=0.65]{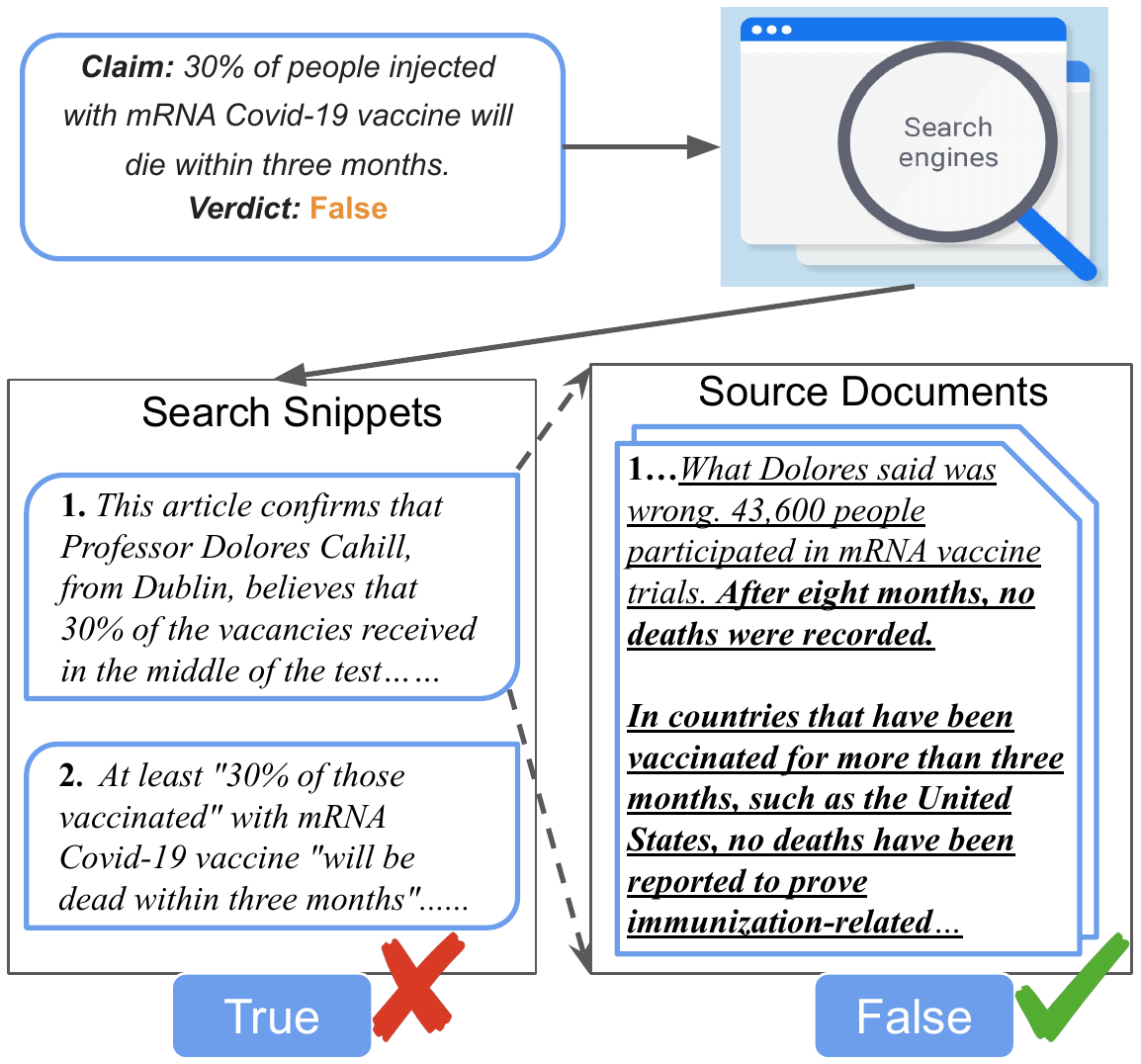}
    \vspace{-2mm}
    \caption{An example claim from the XFact dataset. The example is translated into English for illustration. Search snippet 1 is generated automatically by the search engine, which is a short summary of source document 1. One will predict the claim to be true only based on the search snippets, but the claim is false if the document is provided. }
    \label{fig:xfact}
    \vspace{-3mm}
\end{figure}
\input{tables/dataset}

The spread of misinformation in the modern media ecosystem has become an urgent social issue~\citep{vosoughi2018spread}. In order to combat the proliferation of misleading information, fact-checking becomes an essential task, which aims to assess the factuality of a given claim made in written or spoken language based on the collected evidence~\citep{Adair2017ProgressT, 2018graves}. Figure~\ref{fig:xfact} shows a real-world claim that originates and circulates on Arabic social media. The claim is included in the multilingual dataset XFact~\citep{gupta-srikumar-2021-x}. Here, we translate the claim into English for illustration. In order to evaluate the factuality of this claim, a journalist needs to search through potentially many sources to find the statistics of mRNA vaccines, and perform the numerical comparison based on the evidence.






Though evidence plays a significant role in fact-checking, early efforts in automatic systems only use the claim to predict the factuality~\citep{Rashkin2017TruthOV, turenne2018rumour,jiang2020hover}. \citet{Schuster2020TheLO} demonstrated that relying on surface patterns of claims without considering other evidence fails to identify well-presented misinformation. To address this issue, recent efforts asked annotators to mutate sentences from Wikipedia articles to create claims and evidence~\citep{Thorne2018FEVERAL,jiang2020hover}. These synthetic claims cannot replace real-world claims that are circulating in the media ecosystem as shown in Figure~\ref{fig:xfact}. Therefore, other works chose to crawl real-world claims from fact-checking websites~\citep{Augenstein2019MultiFCAR, gupta-srikumar-2021-x,schlichtkrull2023averitec}, and used the snippets returned by search engines as the evidence. However, such snippets may not provide sufficient information to verify the claim. Take Figure~\ref{fig:xfact} as an example. Based on the snippets only, the verdict of the claim is \textit{True} as necessary information about deaths after vaccination is missing. Through manual inspections, we found that only 46\% of search snippets provide sufficient information, while 82\% of source documents provide ample information to verify the claim. Here document 1 reveals that no deaths have been reported in the vaccine trials and countries that have been vaccinated for more than 3 months. Based on the evidence, one can predict the factuality of the given claim is \textit{False}.






Aiming for improving real-world fact-checking systems, we propose to incorporate full text from source documents as evidence. Unlike previous synthetic datasets, where gold evidence sentences are annotated~\citep{Thorne18Fact,jiang2020hover,vitaminc2021}, the key challenge of using source documents is how to extract related sentences as evidence. Therefore, we are not able to train an evidence extractor in a supervised manner. On the other hand, source documents returned by the search engine contain lots of irrelevant information. Taking such noisy sentences as evidence propagates errors to the downstream claim verification module. In order to address these two issues, we develop a latent variable model, allowing discrete evidence extraction and claim verification in an end-to-end fashion. Our model 
directly controls sparsity and contiguity for maintaining a better balance in keeping relevant sentences as evidence and removing irrelevant sentences. Experiments on two datasets under different settings demonstrate the effectiveness of the proposed approaches. Our key contributions are summarized as follows: 
\begin{itemize}[itemsep=0pt,topsep=0pt,partopsep=0pt]
    \item We conduct extensive analyses on the real-world multilingual dataset XFact, then propose to incorporate source documents and introduce two enriched datasets.
    \item We propose a joint system that models evidence extraction as a latent variable, maintaining a better balance between keeping relevant and removing irrelevant information.
    \item Experiments show that modeling source documents lead to significant improvements upon best-reported models.
\end{itemize}
\vspace{-2mm}

%% file: tables/dataset.tex
\begin{table*}[t!]
\centering
\caption{Comparisons of fact-checking datasets. Type in the header means the type of evidence used, such as sentence (sent), metadata (meta), question-answer pairs (qa pairs), etc. Source means where the evidence is collected from, such as Wikipedia, fact-checking websites. Retrieved denotes if the evidence is given or retrieved from the source.}
\vspace{-2mm}
\scalebox{0.87}{
\begin{tabular}{lcccccccc}
\toprule
\multirow{2}{*}{Dataset} & \multirow{2}{*}{Real-world} & \multirow{2}{*}{Domain} & \multirow{2}{*}{\#Claims} & \multirow{2}{*}{\#Labels}
& \multicolumn{4}{c}{Evidence} \\ 
\cline{6-9}
& & & & & Type & Source &  Retrieved & Avg. Length  \\\midrule
FEVER~\citep{Thorne2018FEVERAL} & \textcolor{orange}{\ding{55}} & Multiple & 185,445 & 3 & Sent & Wikipedia & \textcolor{blue}{\ding{52}} & 1.2 Sents \\
HOVER~\citep{jiang2020hover} & \textcolor{orange}{\ding{55}} & Multiple & 26,171 & 3 & Sent & Wikipedia & \textcolor{blue}{\ding{52}} & 3.1 Sents \\
TabFact~\citep{Chen2020TabFactAL} & \textcolor{orange}{\ding{55}} &  Multiple & 92,283 & 2 & Table & Wikipedia & \textcolor{orange}{\ding{55}} & 1.0 Table \\
InfoTabs~\citep{Gupta2020INFOTABSIO} & \textcolor{orange}{\ding{55}} &  Multiple & 23,738 & English & Table & Wiki & \textcolor{orange}{\ding{55}} & \textcolor{blue}{\ding{52}} \\
ANT~\citep{Khouja2020StancePA} & \textcolor{orange}{\ding{55}} &  Multiple & 4,547 & Arabic  & \textcolor{orange}{\ding{55}} & \textcolor{orange}{\ding{55}} & \textcolor{orange}{\ding{55}} & \textcolor{orange}{\ding{55}} \\
VitaminC~\citep{vitaminc2021} & \textcolor{orange}{\ding{55}} &  Multiple & 488,904 & 3 & Sent & Wikipedia & \textcolor{orange}{\ding{55}} & 1.0 Sents \\
FEVEROUS~\citep{Aly2021FEVEROUSFE} & \textcolor{orange}{\ding{55}} &  Multiple & 87,026 & 3 & Sent/Table & Wikipedia & \textcolor{blue}{\ding{52}} & 1.4 Sents/0.8 Tables \\
\midrule
PunditFact~\citep{Rashkin2017TruthOV} & \textcolor{blue}{\ding{52}} & Multiple & 4,361 & 3 & \textcolor{orange}{\ding{55}} & \textcolor{orange}{\ding{55}} & \textcolor{orange}{\ding{55}} & \textcolor{orange}{\ding{55}} \\
Liar~\citep{Wang2017LiarLP} & \textcolor{blue}{\ding{52}} & Politics & 12,836 & 6 & Meta & Fact-check & \textcolor{orange}{\ding{55}} & \textcolor{orange}{\ding{55}} \\
Snopes~\citep{Hanselowski2019ARA} &  \textcolor{blue}{\ding{52}} & Multiple & 6,422 & 3  & Sent & Fact-check & \textcolor{orange}{\ding{55}} & 5.0 Sents \\
MultiFC~\citep{Augenstein2019MultiFCAR} &  \textcolor{blue}{\ding{52}} & Multiple & 36,534 & 2-27 & Snippet & Internet & \textcolor{blue}{\ding{52}} & 35.7 Sents \\
FakeCovid~\citep{shahifakecovid} &  \textcolor{blue}{\ding{52}} & Health & 5,182 & 2 & \textcolor{orange}{\ding{55}} & \textcolor{orange}{\ding{55}} & \textcolor{orange}{\ding{55}} & \textcolor{orange}{\ding{55}} \\
AnswerFact~\citep{Zhang2020AnswerFactFC} & \textcolor{blue}{\ding{52}} & Product & 60,864 & 5 & Answer & Amazon & \textcolor{orange}{\ding{55}} & 5.0 Sents \\
SciFact~\citep{Wadden2020FactOF} & \textcolor{blue}{\ding{52}} & Science & 1,409 & 3 & Sent & Paper & \textcolor{orange}{\ding{55}} & 1.2 Sents \\
PublicHealth~\citep{kotonya2020health} & \textcolor{blue}{\ding{52}} & Health & 11,832 & 4 & Sent & Fact-check & \textcolor{orange}{\ding{55}} & 5.0 Sents \\
XFact~\citep{gupta-srikumar-2021-x} &  \textcolor{blue}{\ding{52}} & Multiple & 31,189 & 7 & Snippet & Internet & \textcolor{blue}{\ding{52}} & 15.5 Sents \\
WatClaimCheck~\citep{KhanWP22} &  \textcolor{blue}{\ding{52}} & Multiple & 33,697 & 386 & Sent & Fact-check & \textcolor{orange}{\ding{55}} & 64.8 Sents \\
ClaimDecomp~\citep{ChenSCD22} &  \textcolor{blue}{\ding{52}} & Multiple & 1,250 & 6 & QA Pairs & Fact-check & \textcolor{orange}{\ding{55}} & 5.4 Sents \\
AveriTec~\citep{schlichtkrull2023averitec} &  \textcolor{blue}{\ding{52}} & Multiple & 4,568 & 4 & QA Pairs & Internet & \textcolor{blue}{\ding{52}} &  7.1 Sents \\

\bottomrule
\end{tabular}}
\vspace{-4mm}
\label{tab:dataset}
\end{table*}

%% file: subfiles/2_related.tex
\section{Related Work}
\label{related}

\subsection{Fact-Checking Datasets} 

We reviewed the existing fact-checking dataset as summarized in Table~\ref{tab:dataset}. Following~\citet{guo2021}, we grouped the datasets into two categories: real-world and synthetic. Real-world datasets consist of claims that are naturally occurred and fact-checked by journalists, while synthetic datasets contain claims created artificially by mutating sentences from Wikipedia articles. Early real-world efforts predicted the veracity solely based on the claims or with metadata~\citep{Rashkin2017TruthOV, Wang2017LiarLP}, but relying on surface patterns of claims without considering the state of the world fails to identify well-presented misinformation~\citep{Schuster2020TheLO}. Therefore, later works proposed to incorporate evidence into the dataset. \citet{Ferreira2016EmergentAN} used the headlines of selected news articles, and~\citet{pomerleau2017fake} used the entire articles instead as the evidence for the same claims. Instead of using news articles,  \citet{Hanselowski2019ARA} and \citet{kotonya2020health} extracted summaries accompanying fact-checking articles about the claims as evidence. The aforementioned works assume that evidence is given for every claim, which is not conducive to developing systems that need to retrieve evidence from a large knowledge source. In order to integrate the evidence retrieval for better fact-checking, other efforts created claims artificially. \citet{Thorne2018FEVERAL} and \citet{jiang2020hover} considered Wikipedia as the source of evidence and annotated the sentences supporting or refuting each claim. To address this, \citet{Augenstein2019MultiFCAR} and \citet{gupta-srikumar-2021-x} retrieved evidence from the Internet, but the search results were not annotated. Thus, it is possible that irrelevant information is present in the evidence, while information that is necessary for verification is missing. To construct a better evidence-based dataset, we retrieve documents from web pages and select relevant evidence sentences from documents as evidence. Such a design makes the dataset suitable to train fact-checking systems that can extract evidence from web sources and validate real-world claims based on evidence found on the Internet.

Early efforts predicted the veracity solely based on the claims or with metadata~\citep{Rashkin2017TruthOV, Wang2017LiarLP}, but studies show that predictions that do not consider evidence fails to identify misinformation~\citep{Schuster2020TheLO}. Therefore, synthetic  datasets~\citep{Thorne18Fact,jiang2020hover,vitaminc2021} considered Wikipedia as the source of evidence and annotated the sentences from articles as evidence.  
However, these efforts restricted world knowledge to a single source (i.e. Wikipedia), which is not ideal to develop systems that collect evidence from heterogeneous sources. On the other hand, real-world efforts~\citep{Hanselowski2019ARA, kotonya2020health} extracted summaries accompanying fact-checking articles about the claims as evidence. Nonetheless, using fact-checking articles is not realistic, as they are not available during inference. To address this issue, other datasets~\citep{Augenstein2019MultiFCAR, gupta-srikumar-2021-x} included search snippets generated by Google as evidence. Unlike prior real-world efforts, we propose to directly incorporate retrieved documents to provide more information for better verification.

\subsection{Fact-Checking Systems} 
When verifying synthetic claims, systems often operate as a pipeline consisting of an evidence extraction module and a verification module. Relevant articles are first retrieved from the Wikipedia dump by using entity linking, or TF-IDF~\citep{hanselowski-etal-2018-ukp,Thorne18Fact}.  After obtaining the evidence sentences, a textual entailment model is applied for the claim verification~\citep{luken-etal-2018-qed, Nie2019CombiningFE}. Recent systems employ graph-based models to aggregate evidence. This allows the verification of more complex claims where several pieces of information can be combined~\citep{Zhou2019GEARGE, Liu2020KernelGA, Zhong2020ReasoningOS, schlichtkrull2020}. Due to the difficulty of annotating evidence under real-world scenarios, most systems assume the evidence sentences are given, ignoring the challenge of evidence extraction~\citep{Augenstein2019MultiFCAR,kotonya2020health,gupta-srikumar-2021-x}. Different from these methods, our proposal involves treating evidence extraction as a latent variable. This innovative design empowers our system to efficiently gather evidence from various web sources, thereby verifying real-world claims through evidence discovered online.

%% file: subfiles/3_dataset.tex
\begin{figure}[t!]
\centering
\includegraphics[width=0.7\linewidth]{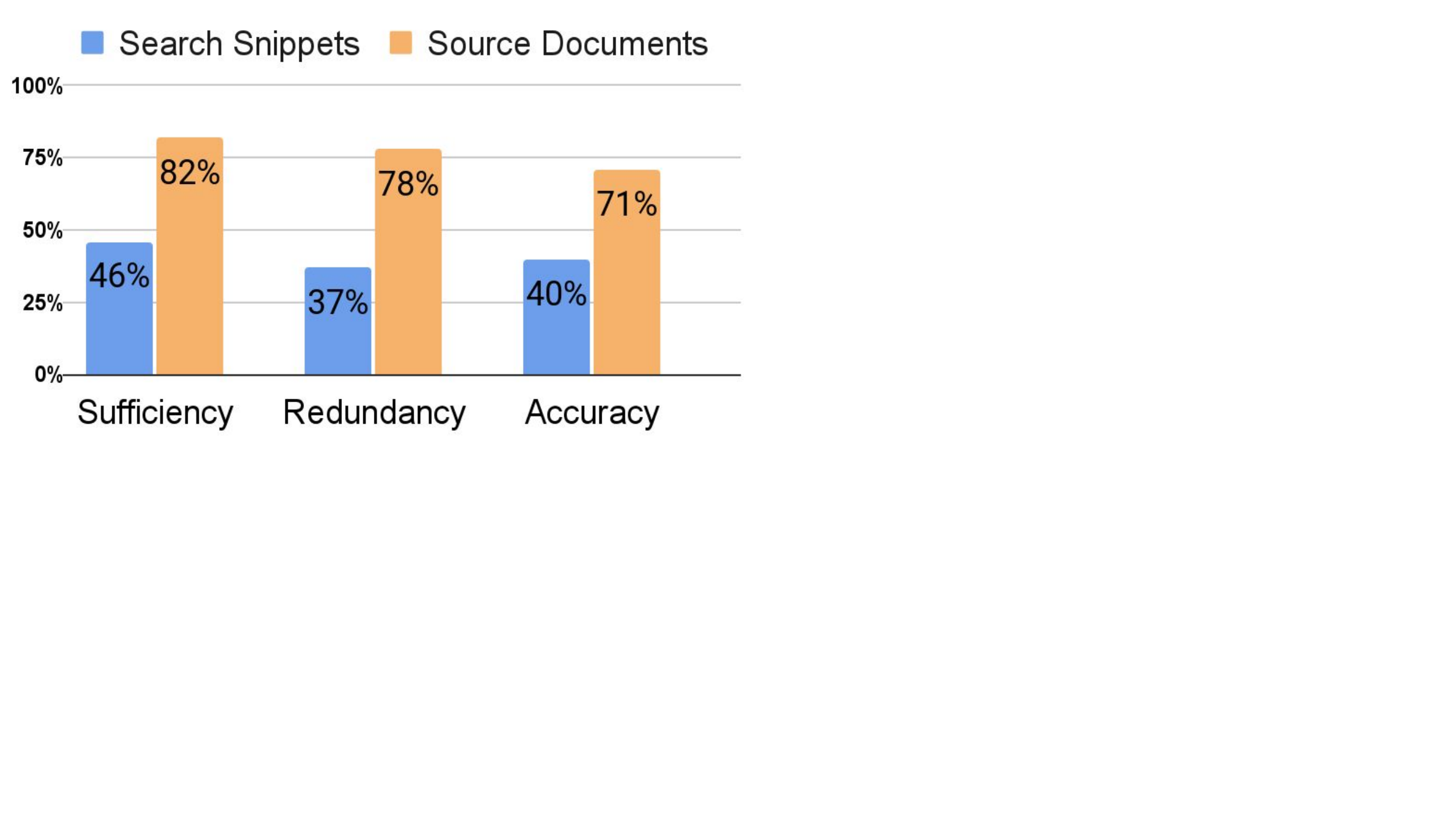}
\vspace{-0.5em}
\caption{Comparison of information sufficiency, redundancy, and prediction accuracy when humans are given search snippets and source documents.}
\label{fig:suf}
\vspace{-5mm}
\end{figure}

\section{Dataset Analysis}
\label{dataset}

\subsection{Search Snippets Analysis}
We investigated the usage of search snippets as evidence for verifying real-world claims. To evaluate the information provided by the search snippets, instances from XFact~\citep{gupta-srikumar-2021-x} were manually examined in two phases. The examination team has fifteen members. Five of them are involved in the first phase, while the other five participants are in the second phase. All annotators are undergraduate students who are fluent in English. To ensure examination consistency, they were trained by the authors and went through several pilot examinations. We randomly selected 100 instances and translated them into English. For inter-annotator agreement, we randomly selected 20\% of claims to be annotated by 5 annotators. We calculated the Fleiss K score~\citep{fleiss1971measuring} to be 0.75.

In the first phase, each annotator was given 50 claims with their corresponding label, search snippets, and source documents. Annotators were required to answer (yes or no) if the snippets and documents provide sufficient information to predict the label of the given claim. We reported the average results in Figure~\ref{fig:suf}, only 46\% of snippets provide sufficient information to verify the claim. Our same analysis suggests that for 82\% of the instances, using documents provides sufficient evidence to determine the factuality. Annotators were also asked to label if each snippet and document is related to the claim. 78\% of source documents are not directly related to the claim (redundancy), while only 37\% of search snippets contain irrelevant information. We also noticed that more than 52\% of the sentences that can be served as evidence were in three consecutive paragraphs in a document. In the second phase, each annotator was given the claim with search snippets and source documents. Annotators were asked to infer the labels of 50 claims based on the snippets or documents. As shown in Figure~\ref{fig:suf}, human predictions are more accurate when source documents are given (71\% vs. 40\%). However, we notice that the performance gap between snippets and documents is smaller in prediction accuracy. One reason is that the documents contain more irrelevant information that may affect the prediction accuracy. Based on these results, we conclude that if a fact-checking system is able to extract relevant sentences from source documents, it will gain benefits from the additional contextual clues.


\input{tables/stats}

\subsection{Multilingual Dataset Extension}
Next, we extend XFact with source documents. XFact contains URLs of web pages and search snippets of these web pages generated by Google search. We only include web pages that the XFact provides the URLs. We also filtered the web pages published after the claims were made to avoid possible information leakage. In detail, we use the HTTP Get method to obtain the web pages according to the given URLs, then utilize xpath to locate all the text content under the <body><p> tags of the web page, so as to exclude the advertisements, brand information, contact information and other irrelevant contents in the web page. Multimedia information (e.g. pictures, videos) is also removed. With this pre-processing procedure, we are able to get textual contents (source documents) from 71\% of the web pages. Due to the difficulty of identifying the evidence sentences from source documents in 25 languages, gold labels of evidence are not annotated. For websites with an anti-crawling function, the above method will not return web pages. In this scenario, we obtain content by manually opening web pages. Table~\ref{tab:split} shows the statistics of the extended XFact. 
\input{tables/distribution}

\subsection{Distribution of the Dataset}
For training and development, the top twelve languages based on the number of labeled examples are included. The average number of examples per language is 1784, with Serbian being the smallest (835). The dataset is split into training
(75\%), development (10\%), and $\alpha_{1}$ test set (15\%). This leaves us with 13 languages for our zero-shot test set (and $\alpha_{3}$). The remaining set of sources form our out-of-domain test set (and $\alpha_{2}$).  In total, X-FACT covers the following 25 languages (shown with their ISO 639-1 code for brevity): ar, az, bn, de, es, fa, fr, gu, hi, id, it, ka, mr, no, nl, pa, pl, pt, ro, ru, si, sr, sq, ta, tr. 

There are 7 possible labels for each claim in XFact and EFact: True, Mostly-True, Partly-True, Mostly False, False, Unverifiable and Other.  Table~\ref{tab:distribution} shows the composition of training, development, and test sets of XFact and EFact, respecitvely.

\subsection{Monolingual Dataset Construction}
In order to comprehensively evaluate the proposed paradigm on real-world claims, we further construct an English dataset EFact. \citet{Augenstein2019MultiFCAR} introduced a real-world English dataset with search snippets as evidence. However, it is constructed 4 years ago, and more than half (58\%) of the URLs provided in the dataset are invalid. The content of the web pages is either deleted or expired, so we are not able to get the texts on the web pages. Following~\citet{gupta-srikumar-2021-x}, we build the monolingual version of XFact. In summary, we scrape fact-checked claims from dedicated agencies and result in a total of 10,000 English claims.  We collected a list of nonpartisan fact-checkers compiled by International Fact-Checking Network (IFCN)\footnote{\url{https://www.poynter.org/ifcn/}}, and Duke Reporter’s Lab\footnote{\url{https://reporterslab.org/fact-checking/}}. After obtaining the list, we first queried Google’s Fact Check Explorer (GFCE)\footnote{\url{https://toolbox.google.com/factcheck/explorer}} for all the fact-checks done by a particular website. Then we crawled the linked article on the website and additional metadata such as author, URL, and date of the claim. We removed duplicate claims and examples where the label appeared in the claim itself. For websites not linked through GFCE, we skipped these websites as the verdict of the claim is not well-specified. 

Next, we normalized the verdict of claims into 7 labels similar to XFact. The label set contains five labels with a decreasing level of truthfulness: True, Mostly-True, Partly-True, Mostly False, and False. To encompass several other cases where assigning a label is difficult due to lack of evidence or subjective interpretations, we introduced Unverifiable as another label. A final label
Other was used to denote cases that do not fall under the above-specified categories. Following the process described, we reviewed each fact-checker’s
rating system along with some examples and manually mapped these labels to our newly designed label scheme. Table~\ref{tab:distribution} shows the label distribution of EFact.

When verifying a claim, journalists first find information related to the fact and evaluate it given the collected evidence. To validate real-world claims, we chose to incorporate full text in the web pages as evidence.
In order to collect evidence from the web sources, we first submitted each claim as a query to the Google Search API by following~\citet{Augenstein2019MultiFCAR} and~\citet{gupta-srikumar-2021-x}. The top 10 search results are retrieved. For each result, we saved the search rank, URL, timestamp and document. For a small percentage of the claims, Google search did not yield any results. We removed these claims from our training, development, and test sets. We have two measures to ensure the reliability of the evidence. Firstly, we maintained a list of misinformation and disinformation websites, all search results from these websites will be filtered out. Secondly, we filtered out results from fact-checking websites to prevent the answer from being trivially found.

%% file: tables/stats.tex
\begin{table}
\centering
\caption{Statistics of XFact and EFact.} 
\vspace{-4mm}
\scalebox{0.92}{
\begin{tabular}{lcccc}
\toprule
Dataset & Type & Train & Dev & Test  \\
\midrule


\multicolumn{1}{l}{\multirow{6}{*}{XFact}}
&

  Claims & 19,079 & 2,535 & 9,106 \\
& Snippets & 85,856 & 11,154 & 47,507  \\
& Documents & 91,579 & 12,089 & 52,552  \\
\cmidrule(lr){2-5}

& \multicolumn{3}{l}{Avg \#Words in the Claim} & 27.8 \\
& \multicolumn{3}{l}{Avg \#Words in the Snippets} & 32.6 \\
& \multicolumn{3}{l}{Avg \#Words in the Documents} & 662.4 \\

\midrule

\multicolumn{1}{l}{\multirow{6}{*}{EFact}}
&

Claims & 8,002 & 801 & 1,197 \\
& Snippets & 44,359 & 4,583 & 6,472  \\
& Documents & 46,524 & 4,650 & 6,954  \\
\cmidrule(lr){2-5}

& \multicolumn{3}{l}{Avg \#Words in the Claim} & 15.6 \\
& \multicolumn{3}{l}{Avg \#Words in the Snippets} & 37.1 \\
& \multicolumn{3}{l}{Avg \#Words in the Documents} & 544.42 \\
\bottomrule
\end{tabular}}
\label{tab:split}
\vspace{-2mm}
\end{table}

%% file: tables/distribution.tex
\begin{table*}[t!]
\centering
\caption{Label distribution of XFact and EFact.}
\vspace{-2mm}
\scalebox{0.9}{
\begin{tabular}{lccccccc}
\toprule
Label & False & Mostly False & Partly True & Mostly True & True & Unverifiable & Other \\
\midrule
Train & 10,836 & 2,872 & 6,565 & 3,589 & 5,716 & 521 & 537 \\
Development & 938 & 105 & 540 & 186 & 521 & 79 & 65 \\
Test (In-Domain) & 1,412 & 148 & 831 & 284 & 797 & 101 & 85 \\
Test (Out-of-Domain) & 980 & 0 & 696 & 0 & 290 & 138 & 12 \\
Test (Zero-Shot) & 1,448 & 411 & 760 & 71 & 327 & 260 & 54  \\
\midrule
Train & 1,549 & 2,077 & 1,834 & 490 & 1,306 & 426 & 320 \\
Development & 155 & 208 & 183 & 49 & 131 & 43 & 32 \\
Test & 232 & 311 & 275 & 73 & 195 & 63 & 48 \\
\bottomrule
\end{tabular}}
\label{tab:distribution}
\vspace{-3mm}
\end{table*}

%% file: subfiles/4_model.tex
\section{Model}
\label{model}

\begin{figure}
    \centering
    \includegraphics[width=\linewidth]{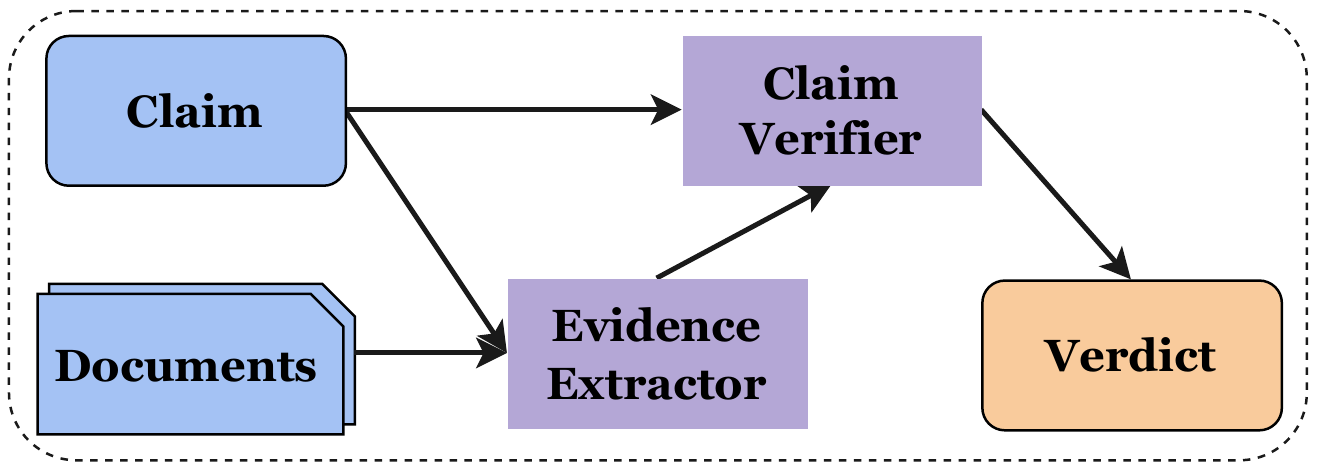}
    \vspace{-4mm}
    \caption{Overview of the model.}
    \vspace{-6mm}
    \label{fig:arch}
\end{figure}

As shown in Figure~\ref{fig:arch}, 
the proposed model contains two modules: evidence extractor and claim verifier. We propose the \textbf{S}parsity and \textbf{C}ontiguity \textbf{A}ssisted \textbf{L}atent \textbf{E}vidence Extractor  (\textbf{SCALE}) as the evidence extractor, which extracts the evidence by assigning binary masks (0 or 1) to sentences. After obtaining the evidence sentences, the claim verifier predicts the verdict of the claim conditioned on the extracted evidence. We provide the pseudo code in Algorithm~\ref{algorithm}. These two modules are jointly trained in an end-to-end manner.

\subsection{Motivation}
There are three main challenges to incorporating source documents for verifying real-world claims. 
\begin{itemize}[itemsep=0pt,topsep=0pt,partopsep=0pt]
    \item The first one is that we cannot train an evidence extractor in a supervised learning manner, as gold evidence sentences in the documents are not available. 
    \item Next, the source documents encompass a substantial amount of irrelevant information, since they are aggregated from heterogeneous web sources.
    \item Lastly, evidence is crucial for generating justifications to convince readers~\citep{guo2021}. Extracted sentences are encouraged to be contiguous, which improves readability~\cite{jain2020learning}. 
\end{itemize}

Aiming at addressing these challenges, we use \textbf{SCALE} to build the model. \textbf{Firstly, we can view the evidence extraction as a latent variable, and jointly train it with claim verification based on SCALE.} Unlike other information retrievers (e.g. TF-IDF), the evidence extractor in the proposed joint system can solicit optimization feedback obtained from the claim verification. On the other hand, \textbf{SCALE} is more stable when compared with other latent variable models~\citep{LeiBJ16,BastingsAT19}, which rely on sampling-based gradient estimators and thus exhibit high variance.

\textbf{Secondly, using SCALE can control sparsity and contiguity in the evidence extraction.} Imposing sparsity helps to strike a balance between removing irrelevant information and keeping relevant information in the document. Encouraging contiguity aids in extracting continuous evidence sentences for better readability.



\vspace{-3mm}
\begin{algorithm}[H]
\small
\caption{Pseudo code implementation of our joint model}
\label{algorithm}
\begin{algorithmic}[1]
\Require Batch size $N$, Claim $c$, Document $d$, Labels $T$
\For {Sampled Mini-batch  $\{c_k\}_{k=1}^N, \{d_k\}_{k=1}^N, \{T_k\}_{k=1}^N$}
\For {All $k \in \{1, ..., N\}$}
\State $C_k, D_k$ = BERT\_encoder($c_k$), BERT\_encoder($d_k$).
\State $Z_k$ = SCALE\_Extractor($C_k$, $D_k$).
\State $T_k'$ = Verifier (concat($C_k, D_k * Z_k$)).  
\State  \Comment{Verifier can be BERT or KGAT}
\EndFor
\State loss = NLLLoss($\{T_k\}_{k=1}^N, \{T'_k\}_{k=1}^N$).
\State loss.backward().
\EndFor\\
\Return Network\{BERT\_encoder, SCALE\_Extractor, Verifier\}.
\end{algorithmic}
\end{algorithm}

\subsection{Evidence Extractor}
We proposed a latent variable model \textbf{SCALE} to compute sentence-level values $ \hat{\mu} \in [0, 1]^L$, which will be used to mask the sentences in the source document. Only the sentences have been assigned non-zero value will be considered as evidence for claim verification.

\paragraph{Sentence Representation}
We use BERT~\citep{Devlin2019BERTPO} to encode the claim and sentences in the document. We feed the sentences independently to the BERT and use the representations of \texttt{CLS} tokens as the sentence representations. Then we concatenate the sentence representations as the claim representations: $\boldsymbol{x}_{c} \in \mathbb{R}^{D {\times}L_{c}}$ and document representations: $\boldsymbol{x}_{d} \in \mathbb{R}^{D {\times}L_{d}}$ where $D$ is the embedding size and $L_{c}$, $L_{d}$ is the number of sentences.

\paragraph{Factor Graph}
Finding the highest-scored evidence sentences under certain constraints can be viewed as a structured prediction problem. Essentially, the global structure can be represented as assignments to multiple variables, and posit a decomposition of the problem into local factors $f$.
Each of $f$ will impose constraints on the evidence sentences. In this work, we introduce two factors: BUDGET and PAIR to control the sparsity and continuity of the extracted sentences. 
In detail, we assume a factor graph $\mathcal{F}$, which consists of each factor $f \in \mathcal{F}$ corresponding to a subset of variables $\boldsymbol{\mu}_{f} = (\boldsymbol{\mu}_{i})_{i \in f}$ in it. Note that $\boldsymbol{\mu}$ is a $L=L_{c}{\times}L_{d}$ dimensional binary mask selecting the evidence sentences from documents that are aligned to the claim.

\begin{figure}
    \centering
    \includegraphics[scale=0.65]{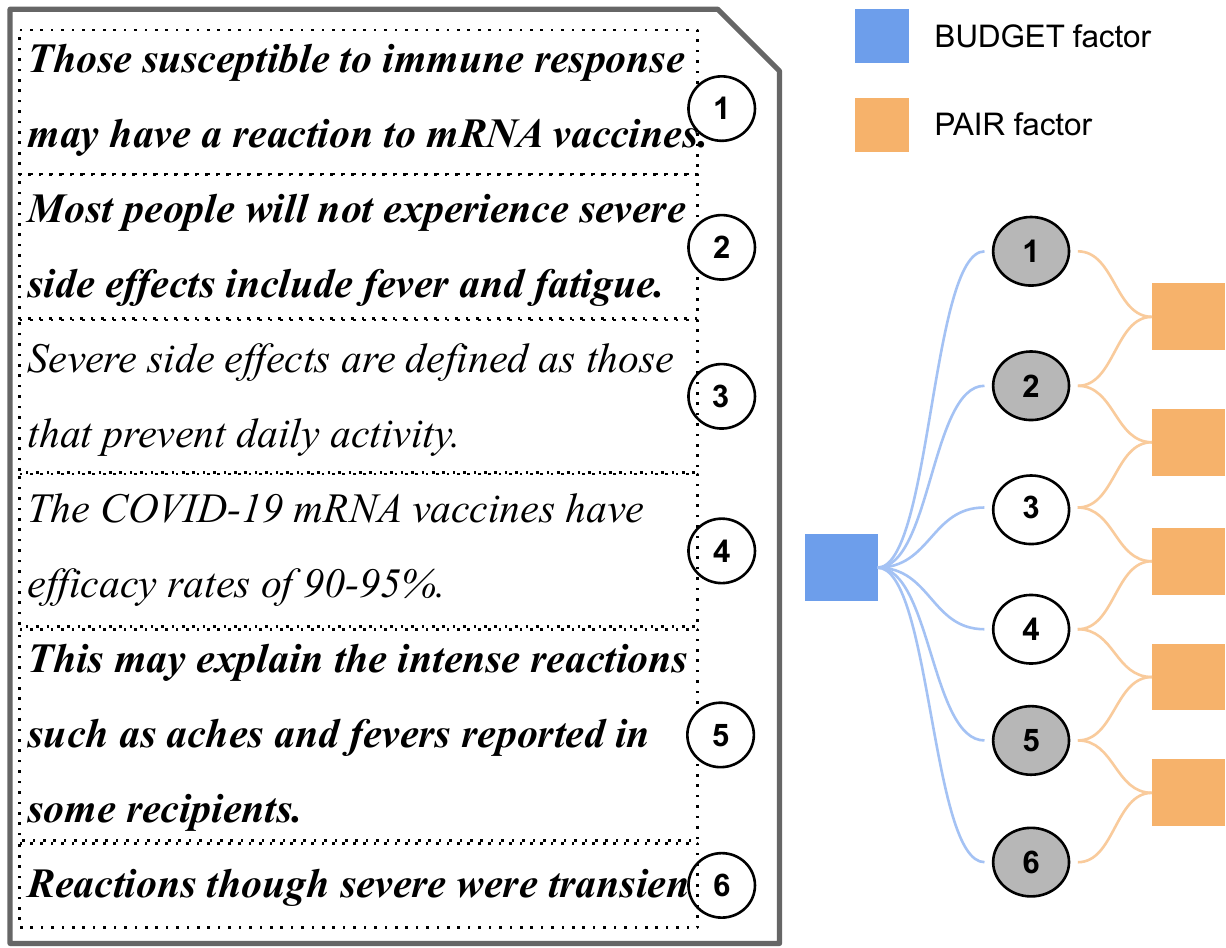}
    \vspace{-0.5em}
    \caption{Factor graph for the evidence extractor.}
    \label{fig:factor}
    \vspace{-5mm}
\end{figure}

The following local sub-problem is required to be tractable for any factor:
\begin{align}\label{local sub-problem}
\hat{\boldsymbol{\mu}}_{f}=\arg \max _{\boldsymbol{\mu}_{f} \in\{0,1\}^{|f|}} \boldsymbol{s}_{f}^{\top} \boldsymbol{\mu}_{f}+h_{f}\left(\boldsymbol{\mu}_{f}\right),
\end{align}
where the first item indicates the selection of sentences in the documents according to the degree of importance matrix: $\boldsymbol{s}_f=\boldsymbol{x}_{c} \cdot \boldsymbol{x}_{d}^{T}$, and the second item represents the local score functions of each factor: $h_f(\boldsymbol{\mu}_f)=\sum_{i=1}^{L-1} r_{i, i+1} \mu_{i, i+1}$, where $r_{i,i+1} \in \mathbb{R}$ are edge scores in factor graph $\mathcal{F}$. Figure~\ref{fig:factor} illustrates how BUDGET factor and PAIR factor are used for imposing constraints on extracting evidence sentences from source documents. PAIR factor is used to impose continuity,
and BUDGET factor is use to induce sparsity. We instantiate a factor graph with $L$ binary variables (one for each sentence) and a pairwise factor for every pair of contiguous sentences:
\begin{align}
    \mathcal{F} = \{PAIR(\mu_{i}, \mu_{i+1}; r_{i, i+1}) : 1 \leq i \leq L \},
\end{align}
A binary pairwise Markov Random Field (MRF) in Equation \ref{local sub-problem} with PAIR factor can be derived as:
\begin{align}
    score (\mu;s) = \sum_{i=1}^{L}s_{i}\mu_{i} + \sum_{i=1}^{L-1} r_{i, i+1} \mu_{i} \mu_{i, i+1},
\end{align}
where $r_{i, i+1} \geq 0$ encourages contiguity on the evidence extraction. We further impose sparsity by adding the BUDGET factor $\sum_{i=1}^{L} \mu_{i}\leq K$ and obtain $\mathcal{F}$ as follows:
\begin{align}
\begin{split}
    \mathcal{F} =
     \{BUDGET(\mu_{1}, ..., \mu_{L}; K) \} \\
     \cup \{PAIR(\mu_{i}, \mu_{i+1}; r_{i, i+1}) : 1 \leq i \leq L \},
\end{split}
\end{align}
where $K$ is a hyperparameter to control the sparsity. More sentences are extracted as $K$ increases.

\paragraph{Marginal Inference} 
To identify the highest-scoring global structure, it is essential to maximize the global score function, denoted as score$(\boldsymbol{\mu}; \boldsymbol{s})$, which combines information coming from all factors.
This can be viewed as the maximum a posteriori (MAP) inference. Formally, it can be written as:
\begin{align}
\label{map}
    \hat{\boldsymbol{\mu}}=\arg \max _{\boldsymbol{\mu} \in\{0,1\}^{L}} \underbrace{\big(\boldsymbol{s}^{\top} \boldsymbol{\mu}+\sum_{f \in \mathcal{F}} h_{f}\left(\boldsymbol{\mu}_{f}\right)\big)}_{\operatorname{score}(\boldsymbol{\mu} ; \boldsymbol{s})}.
\end{align}
The solution to the MAP problem is a vector $\hat{\boldsymbol{\mu}}$ whose entries are 0 and 1. However, it is often difficult to obtain an exact maximization algorithm for complex structured problems that involve interacting sub-problems that have global agreement constraints~\citep{NiculaeM20}. Therefore, we can define a Gibbs distribution $p(\boldsymbol{\mu}; \boldsymbol{s}) \propto exp(score(\boldsymbol{\mu}; \boldsymbol{s}))$. The MAP in Equation~\ref{map} is the mode of this distribution.

\paragraph{SCALE}
Due to the overlapping interaction of the factors $f \in \mathcal{F}$. the MAP problem is often intractable. Continuous relaxation can be used to replace the discrete constraints $\boldsymbol{\mu} \in \{0, 1\}^L$, which is known as LP-MAP inference~\citep{Wainwright2008GraphicalME}. When the factor graph $\mathcal{F}$ does not have cycles, these continuous relaxations are nearly optimal~\citep{KooRCJS10, MartinsFASX15} as for many structured prediction tasks in natural language processing. Formally, we can rewrite the Equation~\ref{map} as:
\begin{align}
\hat{\boldsymbol{\mu}}=\arg \max _{\boldsymbol{\mu} \in[0,1]^{L}} \operatorname{score}(\boldsymbol{\mu} ; \boldsymbol{s}).
\end{align}

However, LP-MAP is not suitable to train with backpropagation. Consequently, we introduce the \textbf{SCALE} model to address the optimization problem. Arbitrary factor graphs can be instantiated as long as a MAP oracle for each factor is provided. In detail, \textbf{SCALE} is the ${l}^2$ regularized LP-MAP as:
\begin{align}
\hat{\boldsymbol{\mu}}=\arg \max _{\boldsymbol{\mu} \in[0,1]^{L}}\left(\operatorname{score}(\boldsymbol{\mu} ; \boldsymbol{s})-1 / 2\|\boldsymbol{\mu}\|^{2}\right).
\end{align}

\input{tables/xfact}

\subsection{Claim Verifier}
Finally, the verifier makes predictions conditioned on the selected evidence and the claim $\boldsymbol{c}$: $\hat{y} = \texttt{pred}(\boldsymbol{\mu}\odot\boldsymbol{x}\mathbin\Vert\boldsymbol{c})$ to obtain the verdict label. $\odot$ and $\mathbin\Vert$ denote the element-wise product and concatenation, respectively. In practice, we adopt two types of claim verifiers as follows: \textbf{(1) BERT-Based Model:}
Following the state-of-the-art model on XFact~\citep{gupta-srikumar-2021-x}, we use a multi-layer perceptron with embeddings from BERT~\citep{Devlin2019BERTPO} to predict the verdict of the claim. \textbf{(2) Graph-Based Model:}
Kernel graph attention network~\citep{Liu2020KernelGA} is the SOTA graph-based verifier on FEVER~\citep{Thorne18Fact}. Following~\citet{Liu2020KernelGA}, we construct the evidence graph based on the output embeddings of the claim and selected evidence. Node and edge kernels are then used to conduct fine-grained evidence propagation. The updated node representations are used to predict the verdict of the claim.


%% file: tables/xfact.tex
\begin{table*}
\centering
\caption{Results on three different test sets of XFact with search snippets (Snip), extra sentences surrounding the snippets (Snip+) and source documents (Doc). $^\dagger$ denotes results from \citet{gupta-srikumar-2021-x}.}
\vspace{-4mm}
\scalebox{0.9}{
\begin{tabular}{ccccccccc}
\toprule
\multicolumn{3}{c}{\multirow{2}{*}{Extractors / Verifiers}} & \multicolumn{3}{c}{BERT-Based Model}  & \multicolumn{3}{c}{Graph-Based Model} \\
\cmidrule(lr){4-6} \cmidrule(lr){7-9}
& & & In-Domain & Out-of-Domain & Zero-Shot & In-Domain & Out-of-Domain & Zero-Shot \\
\midrule

\multicolumn{3}{c}{\multirow{1}{*}{Majority}}
 & 6.90$^\dagger$ & 10.60$^\dagger$  & 7.60$^\dagger$ &  - & - &  -  \\

\midrule

\multicolumn{1}{c}{\multirow{8}{*}{Snip}}
& \multicolumn{1}{|c}{\multirow{8}{*}{Joint}} 

& \multicolumn{1}{|l|}{\multirow{1}{*}{Atten~\cite{gupta-srikumar-2021-x}}} & 38.90$^\dagger$ & 15.70$^\dagger$  & 16.50$^\dagger$ &  39.43\small±1.24 & 16.04\small±0.94  & 16.88\small±1.04   \\
& \multicolumn{1}{|l}{} & \multicolumn{1}{|l|}{Reinforce~\cite{LeiBJ16} } &  39.18\small±1.37 & 17.25\small±1.42  & 17.66\small±1.45  & 39.46\small±1.35 & 17.40\small±1.27  & 17.92\small±1.34 \\
& \multicolumn{1}{|l}{} & \multicolumn{1}{|l|}{FusedMax~\cite{NiculaeB17}} & 38.24\small±1.34 & 16.82\small±1.31  & 17.04\small±1.58  & 38.41\small±1.52 & 17.08\small±1.24  & 17.31\small±1.17 \\
& \multicolumn{1}{|l}{} & \multicolumn{1}{|l|}{Gumbel~\cite{maddisonMT17}} & 
38.31\small±1.28 &	16.61\small±0.86 &	17.11\small±1.05 &	38.55\small±1.34 &	16.82\small±0.95 &	17.33\small±1.18 \\
& \multicolumn{1}{|l}{} & \multicolumn{1}{|l|}{HardKuma~\cite{BastingsAT19}} & 
38.26\small±1.13 &	16.78\small±1.49 &	17.23\small±1.06 &	38.42\small±0.77 &	16.94\small±1.28 &	17.44\small±0.93 \\
& \multicolumn{1}{|l}{} & \multicolumn{1}{|l|}{UNIREX~\cite{chan2022unirex}} & 
38.47\small±1.01 &	16.98\small±0.84 &	17.47\small±1.05 &	38.77\small±0.82 &	17.02\small±0.73 &	17.64\small±1.19 \\
& \multicolumn{1}{|l}{} & \multicolumn{1}{|l|}{DAR~\cite{liu2024enhancing}} & 
38.76\small±1.24 &	17.24\small±0.82 &	17.88\small±1.19 &	38.91\small±0.95 &	17.13\small±0.73 &	17.62\small±1.01 \\
& \multicolumn{1}{|l}{} & \multicolumn{1}{|l|}{\textbf{Ours}} & \textbf{40.88\small±1.14} & \textbf{18.46\small±0.93}  & \textbf{18.73\small±1.21} & \textbf{41.21\small±1.24 } & \textbf{18.79\small±1.03}  & \textbf{19.04\small±1.08} \\

\midrule
\multicolumn{1}{c}{\multirow{2}{*}{Snip+}}
& \multicolumn{1}{|c}{\multirow{2}{*}{Pipe}} 

& \multicolumn{1}{|l|}{\multirow{1}{*}{Rule (6 Sentences)}} & 41.73\small±1.19 &	19.04\small±1.08 &	19.05\small±1.63 &	42.02\small±1.33 &	19.57\small±1.37 &	19.21\small±1.29 \\
& \multicolumn{1}{|l}{} & \multicolumn{1}{|l|}{Rule (12 Sentences)} & 41.57\small±1.37 &	18.89\small±1.42 &	18.73\small±1.51 &	41.83\small±1.40 &	18.97\small±1.29 &	18.79\small±1.38 \\


\midrule

\multicolumn{1}{c}{\multirow{15}{*}{Doc}} & \multicolumn{1}{|c}{\multirow{7}{*}{Pipe}} & \multicolumn{1}{|l|}{\multirow{1}{*}{Surface~\cite{Aly2021FEVEROUSFE}}} & 
42.76\small±1.36 &	19.97\small±1.38 &	20.02\small±1.52 &	42.88\small±1.44 &	20.15\small±1.29 &	20.26\small±1.22 \\
& \multicolumn{1}{|l}{} & \multicolumn{1}{|l|}{\multirow{1}{*}{Semantic~\cite{Nie2019CombiningFE}}} & 
42.89\small±1.42 &	20.04\small±1.58 &	20.10\small±1.45 &	43.01\small±1.37 &	20.22\small±1.28 &	20.43\small±1.21 \\
& \multicolumn{1}{|l}{} & \multicolumn{1}{|l|}{\multirow{1}{*}{Hybrid~\cite{Shaar2020ThatIA}}} & 42.98\small±1.28 &	20.12\small±1.45 &	20.21\small±1.47 &	43.12\small±1.35 &	20.31\small±1.39 &	20.45\small±1.13 \\
& \multicolumn{1}{|l}{} & \multicolumn{1}{|l|}{\multirow{1}{*}{CONCRETE~\cite{huang2022concrete}}} & 
43.27\small±1.06 &	21.33\small±1.14 &	21.84\small±1.01 &	43.41\small±0.77 &	21.52\small±0.93 &	22.03\small±0.93 \\
& \multicolumn{1}{|l}{} & \multicolumn{1}{|l|}{\multirow{1}{*}{CofCED~\cite{yang2022coarse}}} &
43.36\small±1.15 &	21.18\small±0.93 &	21.46\small±1.21 &	43.61\small±0.89 &	21.36\small±1.08 &	21.67\small±0.96 \\
& \multicolumn{1}{|l}{} & \multicolumn{1}{|l|}{\multirow{1}{*}{FDHN~\cite{xu2023fuzzy}}} &
43.12\small±1.03 &	21.02\small±1.22 &	21.09\small±1.08 &	43.28\small±1.11 &	21.24\small±1.21 &	21.01\small±1.05 \\
& \multicolumn{1}{|l}{} & \multicolumn{1}{|l|}{\multirow{1}{*}{GETRAL~\cite{wu2023adversarial}}} &
43.46\small±1.02 &	21.24\small±0.86 &	21.53\small±1.12 &	43.56\small±0.93 &	21.44\small±1.00 &	21.89\small±0.90 \\
\cmidrule(lr){2-9}

& \multicolumn{1}{|c}{\multirow{8}{*}{Joint}} & \multicolumn{1}{|l|}{\multirow{1}{*}{Atten \cite{gupta-srikumar-2021-x}}} & 
43.98\small±0.72 &	21.53\small±0.58 &	21.78\small±0.59 &	44.11\small±0.46 &	21.74\small±0.63 &	22.01\small±0.55 \\
& \multicolumn{1}{|l}{} & \multicolumn{1}{|l|}{Reinforce~\cite{LeiBJ16}} & 
44.12\small±1.38 &	21.86\small±1.28 &	21.99\small±1.52 &	44.34\small±1.34 &	21.92\small±1.42 &	22.13\small±1.35\\
& \multicolumn{1}{|l}{} & \multicolumn{1}{|l|}{FusedMax~\cite{NiculaeB17}} & 
43.78\small±1.48 &	21.46\small±1.29 &	21.87\small±1.55 &	44.15\small±1.43 &	21.82\small±1.61 &	22.10\small±1.52 \\
& \multicolumn{1}{|l}{} & \multicolumn{1}{|l|}{Gumbel~\cite{maddisonMT17}} & 
43.69\small±1.26 & 	21.70\small±1.50 & 	22.04\small±1.41 & 	44.22\small±1.38 & 	21.98\small±1.04 & 	22.24\small±1.44 \\
& \multicolumn{1}{|l}{} & \multicolumn{1}{|l|}{HardKuma~\cite{BastingsAT19}} & 
43.86\small±1.20 &	21.59\small±1.27 &	21.79\small±0.97 &	44.09\small±1.52 &	22.06\small±0.88 &	22.18\small±1.37 \\
& \multicolumn{1}{|l}{} & \multicolumn{1}{|l|}{UNIREX~\cite{chan2022unirex}} & 
44.18\small±0.91 &	21.76\small±1.45 &	22.42\small±1.29 &	44.53\small±0.80 &	22.47\small±0.96 &	22.82\small±1.18 \\
& \multicolumn{1}{|l}{} & \multicolumn{1}{|l|}{DAR~\cite{liu2024enhancing}} & 
44.42\small±0.98 &	21.99\small±1.28 &	22.36\small±1.05 &	44.80\small±0.91 &	22.61\small±0.88 &	23.00\small±1.01 \\

& \multicolumn{1}{|l}{} & \multicolumn{1}{|l|}{\textbf{Ours}} & \textbf{46.04\small±0.83} & \textbf{23.77\small±1.13}  & \textbf{24.38\small±1.06} & \textbf{46.36\small±1.32} & \textbf{24.08\small±0.94}  & \textbf{24.79\small±1.15} \\

\bottomrule
\end{tabular}}
\label{tab:results}
\vspace{-4mm}
\end{table*}

%% file: subfiles/5_experiment.tex
\section{Experiments and Analyses}
\label{sec:experiments}


\subsection{Baselines}
We adopt seven representative extractor baselines. Pipeline extractors select sentences without seeking supervision from the verdict prediction. Joint extractors train evidence extraction and claim verification jointly. As an indicator of label distribution, we include a majority baseline with the most frequent label of the distribution.

\subsubsection{Pipeline Extractors}

\noindent \textbf{(1) Rule-based Extractor}: This is a simple baseline that includes the $N$ sentences adjacent the snippet in the source document. In practice, we choose $N=6$ and $N=12$. 

\noindent \textbf{(2) Surface Extractor}: Following previous efforts on synthetic datasets~\citep{Thorne2018FEVERAL,jiang2020hover,Aly2021FEVEROUSFE}, we use TF-IDF to extract sentences in the source documents as the evidence.

\noindent \textbf{(3) Semantic Extractor}:  
Following~\citet{Nie2019CombiningFE}, we extract evidence based on semantic similarity. BERT~\citep{Devlin2019BERTPO} is used to get the representations of the claim and sentences in the source document. Cosine similarity is used for selecting evidence.

\noindent \textbf{(4) Hybrid Extractor}: We employ rankSVM to choose sentences based on the feature sets of rankings returned by TF-IDF as well as similarity scores calculated using BERT.

\noindent \textbf{(5) CONCRETE}: \citet{huang2022concrete} introduces a pioneering fact-checking framework utilizing cross-lingual retrieval, gathering evidence from various languages using a retriever.

\noindent \textbf{(6) CofCED}: \citet{yang2022coarse} employs a hierarchical encoder for web text representation, develops cascaded selectors for verdict-explainable sentence selection.

\noindent \textbf{(7) FDHN}: \citet{xu2023fuzzy} presents a fuzzy logic-based hybrid model that combines deep learning with textual and numerical context analysis to enhance fake news detection.

\noindent \textbf{(8) GETRAL}: \citet{wu2023adversarial} models claims and evidences as graph-structured data, focusing on capturing long-distance semantic dependencies through neighborhood propagation.

\subsubsection{Joint Extractors}

\noindent \textbf {(1) Attention}: Following~\citet{gupta-srikumar-2021-x}, we get relevance weights between the output embeddings of all retrieved sentences and the claim via dot product attention. Then we obtain the evidence by filtering the weighted sentences.

\noindent \textbf{(2) Reinforce}: We follow~\citet{LeiBJ16} by assigning a binary Bernoulli variable to each sentence from source documents. The evidence extractor is optimized using REINFORCE~\citep{williams1992simple}. A $L_{0}$ regularizer is used to impose sparsity.

\noindent \textbf{(3) FusedMax}: We used fusedmax~\citep{NiculaeB17} to encourage attention to contiguous segments of text, by adding an additional total variation regularizer, inspired by the fused lasso.

\noindent \textbf{(4) Gumbel}: Following~\citet{maddisonMT17}, we employ the Gumbel-Max trick to reparameterize the Bernoulli variables.

\noindent \textbf{(5) HardKuma}: We follow~\citet{BastingsAT19} by adopting HardKuma variables and use reparameterized gradients estimates~\citep{KingmaW14}.

\noindent \textbf{(6) UNIREX}: We adopt the rationale extraction in \citet{chan2022unirex}.

\noindent \textbf{(7) DAR}: Discriminatively Aligned Rationalization (DAR) \cite{liu2024enhancing} uses an auxiliary module to ensure alignment between the selected rationale and the original input.



\subsection{Experimental Settings}
Following~\citet{gupta-srikumar-2021-x}, we evaluate our proposed model on three test sets. The out-of-domain and zero-shot test sets aim to measure the transfer abilities of a fact-checking system across different domains and languages.
\textbf{(1) In-Domain}: The test set ($\alpha_{1}$) is distributionally similar to the training set, and contains claims from the same languages and sources as the training set.
\textbf{(2) Out-of-Domain}: The test set ($\alpha_{2}$) contains claims from the same languages as the training set but from different domains. A model performs well on both $\alpha_{1}$ and $\alpha_{2}$ can generalize across different domains.  
\textbf{(3) Zero-Shot}: The test set ($\alpha_{3}$)  includes claims from languages not contained in the training set. Models that overfit language-specific artifacts will have poor performance on $\alpha_{3}$. 

We reported the mean F1 score and standard deviation by 5 runs.

\paragraph{Hyperparameters of the pipeline and joint extractors.}\label{Hyperparameters}
Following previous efforts on synthetic datasets~\citep{Thorne2018FEVERAL,jiang2020hover}, we configure the pipeline extractor to select five pieces of evidence from source documents. For the pipeline extractors, we set the retrieved evidence obtained from TF-IDF to be more than $5$ words for surface extractor. We use the mBERT default tokenizer with max-length as $256$ to preprocess data for semantic extractor. We use the default parameters in scikit-learn with RBF kernel for the hybrid extractor. For the joint extractors, we build the models based on their official implementations and tune the hyper-parameters on the dev set. For the proposed model, we use Adam~\citep{kingma2014adam} with $1e{-5}$ learning rate with $0.5$ decay. The predictor hidden size is set to $200$.

Documents from specialized domains such as science and e-commerce have also been considered~\citep{Wadden2020FactOF, Zhang2020AnswerFactFC}. 
\citet{vitaminc2021} constructed VitaminC based on factual revisions to Wikipedia, in which evidence pairs are nearly identical in language and content, with the exception that one supports a claim while the other does not. However, these efforts restricted world knowledge to a single source (Wikipedia), ignoring the challenge of retrieving evidence from heterogeneous sources on the web.

\vspace{-4mm}
\subsection{Main Results}
\paragraph{XFact:} Table~\ref{tab:results} shows our model surpassing others on three tests using search snippets or source documents. It also generalizes across domains ($\alpha_{2}$) or languages ($\alpha_{3}$) with robust standard deviation. Providing more context around snippets enhances results. Yet, adding more sentences results in a performance improvement of less than 1\%. The enhancement in the model's performance is not always due to the potential non-adjacency of evidence sentences.

\input{tables/efact}


Joint extractors using source documents perform better than snippet models, with our model showing a 5.39\% improvement, highlighting the value of more context. The graph-based model has a higher F1 score than the BERT-based one, as real-world claim verification demands multiple evidence synthesis. Comparing extractors with source documents, hybrid extractors outperform both surface and semantic ones in pipeline methods. Joint approaches beat all pipeline ones, our model has improved by an average of 2.68\% compared to the pipeline model's state-of-the-art, GETRAL, emphasizing the value of more context. emphasizing the significance of joint training for evidence extraction and factuality prediction. This joint training provides explicit feedback and greater robustness in terms of standard deviation. The inconsistency in pipeline extractors' results is due to excess irrelevant data. Our model surpasses other joint models in performance and robustness, showing a 1.82\% improvement over the SOTA joint model, DAR. This is primarily due to \textbf{SCALE} allowing for deterministic and fully differentiable evidence extraction, resulting in a sturdy and well-generalized model.



\begin{figure}
    \centering
    \includegraphics[scale=0.4]{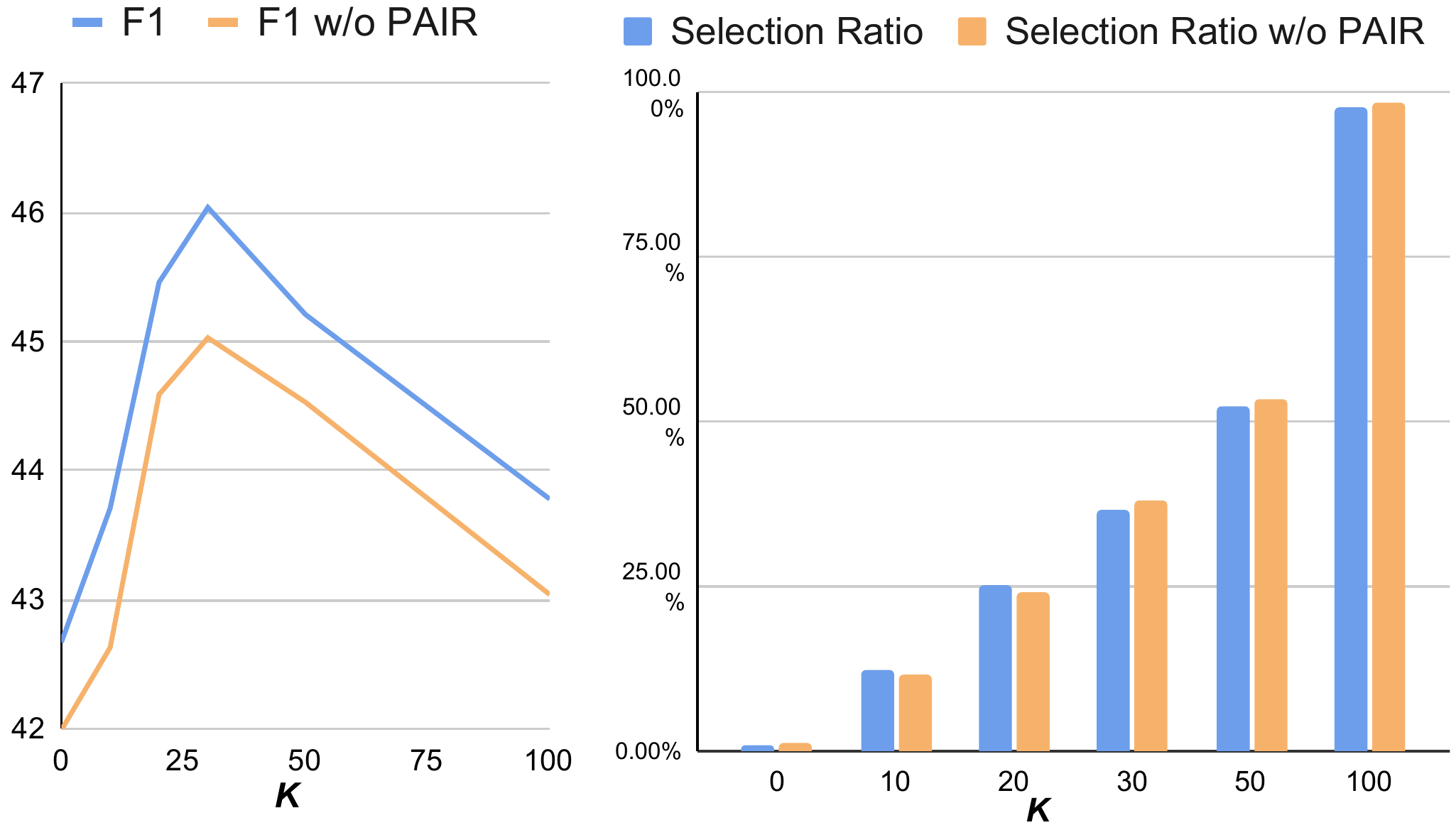}
    \vspace{-0.5em}
    \caption{Effects of Factors (BUDGET and PAIR). BUDGET is imposed to control the sparsity of the sentence selection. $K$ is the hyper-parameter to control it. PAIR is imposed to encourage contiguity. }
    \label{fig:factorK}
    \vspace{-4mm}
\end{figure}

\paragraph{EFact:} 
We further present the experimental results on the monolingual dataset in Table~\ref{tab:Efactresults}. We observe similar experimental results as for XFact. Our proposed model consistently outperforms baselines under different settings. Given the search snippets, our model outperforms other joint models by 0.88\% on average. When the source documents are provided, the performance gap becomes larger (2.6\% on average). Such results further illustrate the effectiveness of incorporating source documents to verify real-world claims and the superiority of the proposed model when more contexts are given. 



\subsection{Analysis and Discussion}
In this section, we give detailed analyses for baselines on XFact with snippets and documents. 

\paragraph{Effect of Factors:}
As shown in Figure~\ref{fig:factorK}, as $K$ increases, more sentences in the source documents are selected as evidence. However, the F1 score of the model is not monotonically increasing as $K$ increases, as irrelevant information is included. The model achieves the best performance when $K=30$, where 11.74 sentences (191.07 tokens) are selected as evidence on average. If we remove the BUDGET factor completely, the performance will drop from 46.04 to 45.03 in terms of the F1 score. On the other hand, the PAIR constraint is also beneficial to the model. Models with PAIR constraints consistently achieve better results.

\paragraph{Effect of Metadata:}
We further study the impact of concatenating metadata (e.g. language, publish date) to the claim for verification. From Table~\ref{tab:meta}, the proposed model gains performance improvements from the metadata. However, improvements in the model with documents are smaller when compared with the model that uses snippets. Also, the impact of metadata is less significant than the ratio of extracted sentences. The model without metadata can achieve competitive results than models with metadata but less optimal extraction ratio. One potential reason is that the model can extract important information similar to metadata from documents, so the impact of metadata is diminished.



\input{tables/metadata}

\input{tables/evidence_number}

\input{tables/human}
\begin{figure*}
    \centering
    \includegraphics[width=0.96\linewidth]{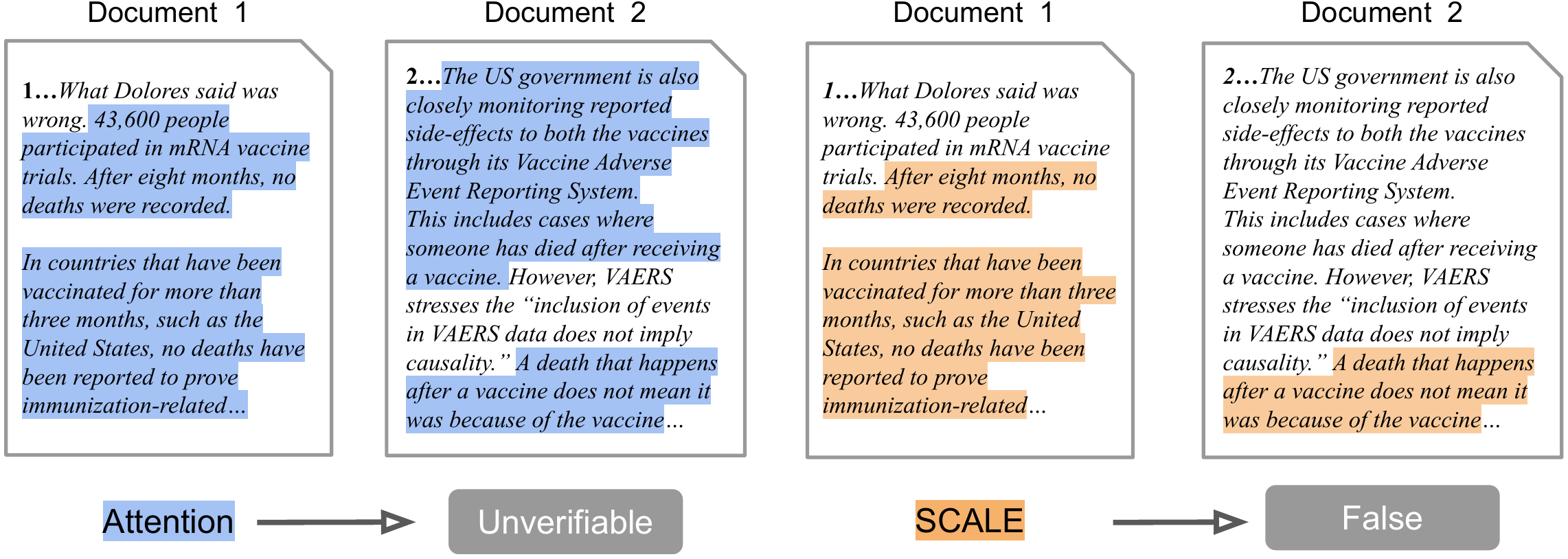}
    \caption{A case on extracted sentences from source documents based on attention and \textbf{SCALE}.}
    \vspace{-4mm}
    \label{fig:case}
\end{figure*}
\begin{figure}
    \centering
    \includegraphics[width=0.99\linewidth]{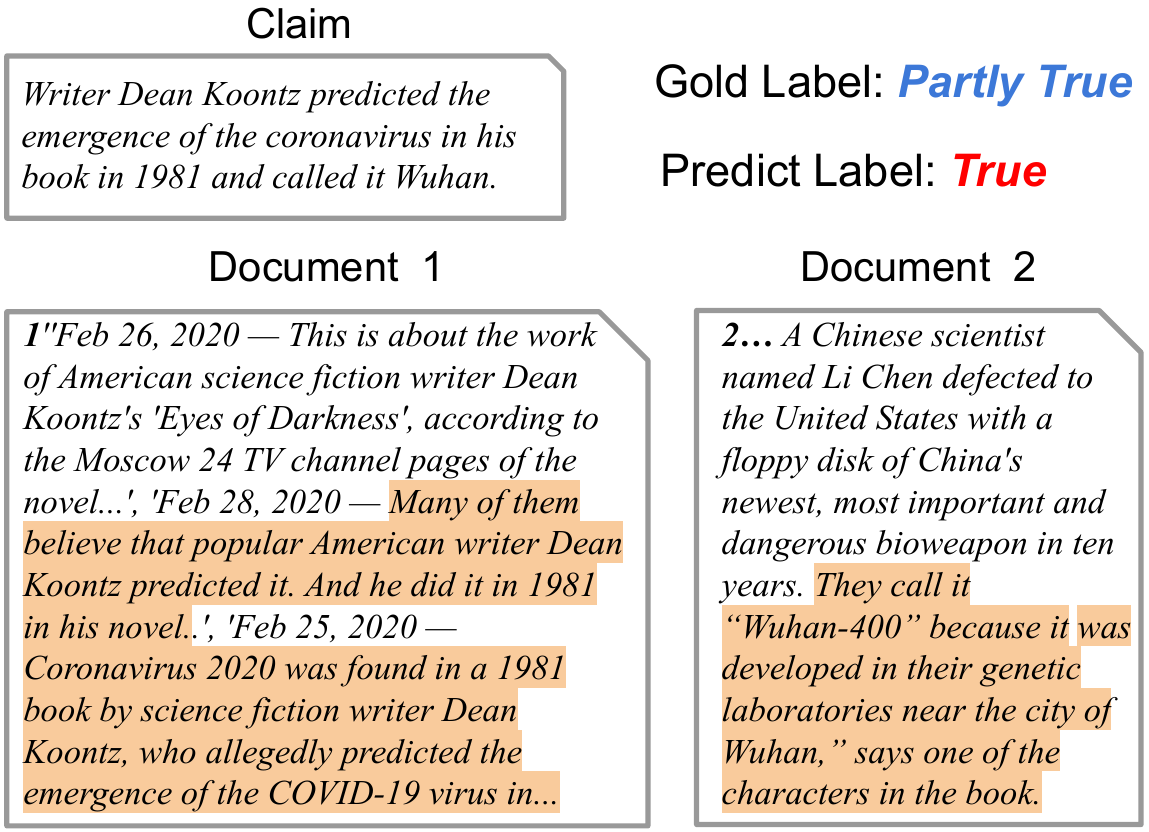}
    \caption{An error case on extracted sentences from source documents based on \textbf{SCALE}.}
    \vspace{-4mm}
    \label{fig:error case}
\end{figure}

\paragraph{Effect of Evidence:}
In Table~\ref{tab:evidence}, we vary the number of evidence from source documents for pipeline extractors and report the F1 scores on the test set with the BERT-based model. The variation results indicate that both the quantity and quality of retrieved evidence affect the performance. Using less evidence could not provide enough information to help the verifiers to predict the factual label. In contrast, introducing too much evidence will bring irrelevant and noisy sentences thus impeding the veracity prediction.

\paragraph{Human Evaluaion:}
We asked 5 annotators to annotate 100 claims sampled from the test set $\alpha_{1}$. Since XFact is a multilingual dataset, we first translated these claims and source documents into English. Each annotator is required to select sentences that would be able to verify the claim as evidence. We compare sentences extracted by different extractors including surface, semantic, and hybrid extractors with different numbers of sentences with the gold evidence annotated by the annotators, and show the results in Table \ref{tab:ex}. Compared with the other baseline model, our method can obtain a 7.98\% F1 performance boost on average. When compared with pipeline extractors, our model can maintain a better balance between precision and recall. Such results show the effectiveness of our approach that jointly models evidence extraction and claim verification.

\paragraph{Case Study:}
We present the case study of extracting evidence on source documents using joint extractors based on attention and \textbf{SCALE}. From Figure \ref{fig:case}, we observe that attention based joint extractor extract much more evidence on source documents than \textbf{SCALE} based extractor, thus introducing more irrelevant information and making it more difficult for verifiers to predict factual labels. We attribute the accurate and effective evidence obtained by \textbf{SCALE} to constrained (e.g. sparsity, contiguity), deterministic and fully differentiable extracting capabilities.
\paragraph{Error Analysis:}
We present a typical error case made by the proposed model on the out-of-domain set as shown in Figure~\ref{fig:error case}. The gold label of this claim is Partly True, while the model predicted it as True. The extracted sentences shows the model does able to find relevant evidence to support the claim. However, the claim verifier has difficulties in predicting labels with similar ratings (partly true v.s. true). One reason is that the claim verifier is not trained on this domain, which makes it harder to distinguish labels with similar ratings. Through calculation, we found that more than 80\% of the errors are caused by predicting similar labels within two ratings. For example, predicting mostly false and partly true as false. On the in-domain test set, the proposed model made 24\% less such errors. Another reason is that distinguishing similar ratings is a very difficult problem. Even for human fact-checkers, their ratings on the same claim are not consistent~\citep{guo2021}. The training corpus potentially exhibit such inconsistency, which makes the model confused when predicting the factuality.




%% file: tables/efact.tex
\begin{table}[t!]
\centering
\caption{Results on test set of EFact with search snippets (Snip), extra sentences surrounding the snippets (Snip+) and source documents (Doc).}
\vspace{-4mm}
\scalebox{0.84}{
\begin{tabular}{ccccc}
\toprule
\multicolumn{3}{c}{Extractors / Verifiers} & \multicolumn{1}{c}{BERT}  & \multicolumn{1}{c}{Graph} \\
\midrule

\multicolumn{3}{c}{\multirow{1}{*}{Majority}}
 & 7.58 & -   \\

\midrule

\multicolumn{1}{c}{\multirow{8}{*}{Snip}}
& \multicolumn{1}{|c}{\multirow{8}{*}{Joint}} 

& \multicolumn{1}{|l|}{\multirow{1}{*}{Atten}} & 36.83$^\dagger$ & 36.75$^\dagger$    \\
& \multicolumn{1}{|l}{} & \multicolumn{1}{|l|}{Reinforce } &  37.58\small±1.46 & 37.84\small±1.45  \\
& \multicolumn{1}{|l}{} & \multicolumn{1}{|l|}{FusedMax} & 36.53\small±0.87 & 36.83\small±0.81  \\
& \multicolumn{1}{|l}{} & \multicolumn{1}{|l|}{Gumbel} & 36.29\small±0.93 & 35.70\small±0.80  \\
& \multicolumn{1}{|l}{} & \multicolumn{1}{|l|}{HardKuma} & 37.49\small±1.41 & 37.39\small±1.30  \\
& \multicolumn{1}{|l}{} & \multicolumn{1}{|l|}{UNIREX} & 37.65\small±1.24 & 37.44\small±1.35  \\
& \multicolumn{1}{|l}{} & \multicolumn{1}{|l|}{DAR} & 37.52\small±1.27 & 37.41\small±1.28  \\

& \multicolumn{1}{|l}{} & \multicolumn{1}{|l|}{\textbf{Ours}} & \textbf{38.09\small±1.20} & \textbf{37.92\small±1.31}  \\

\midrule
\multicolumn{1}{c}{\multirow{2}{*}{Snip+}}
& \multicolumn{1}{|c}{\multirow{2}{*}{Pipe}} 

& \multicolumn{1}{|l|}{\multirow{1}{*}{Rule (6 Sents)}} & 39.66\small±1.48 & 39.76\small±1.56    \\
& \multicolumn{1}{|l}{} & \multicolumn{1}{|l|}{Rule (12 Sents)} & 38.42\small±1.01 & 38.11\small±0.88 \\

\midrule

\multicolumn{1}{c}{\multirow{11}{*}{Doc}} & \multicolumn{1}{|c}{\multirow{3}{*}{Pipe}} & \multicolumn{1}{|l|}{\multirow{1}{*}{Surface}} & 39.85\small±1.62 & 40.02\small±1.17   \\
& \multicolumn{1}{|l}{} & \multicolumn{1}{|l|}{\multirow{1}{*}{Semantic}} & 40.35\small±1.68 & 40.42\small±1.34  \\
& \multicolumn{1}{|l}{} & \multicolumn{1}{|l|}{\multirow{1}{*}{Hybrid}} & 40.27\small±1.10 & 40.58\small±1.35   \\
\cmidrule(lr){2-5}

& \multicolumn{1}{|c}{\multirow{8}{*}{Joint}} & \multicolumn{1}{|l|}{\multirow{1}{*}{Atten}} & 40.13\small±0.92 & 20.04\small±0.65   \\ 
& \multicolumn{1}{|l}{} & \multicolumn{1}{|l|}{Reinforce} & 41.65\small±0.95 & 40.95\small±1.18   \\
& \multicolumn{1}{|l}{} & \multicolumn{1}{|l|}{FusedMax} & 42.39\small±1.18 & 42.03\small±1.12  \\
& \multicolumn{1}{|l}{} & \multicolumn{1}{|l|}{Gumbel} & 42.12\small±1.45 & 42.32\small±1.19  \\
& \multicolumn{1}{|l}{} & \multicolumn{1}{|l|}{HardKuma} & 43.38\small±1.11 & 43.55\small±1.16 \\
& \multicolumn{1}{|l}{} & \multicolumn{1}{|l|}{UNIREX} & 43.46\small±1.24 & 43.72\small±1.05 \\
& \multicolumn{1}{|l}{} & \multicolumn{1}{|l|}{DAR} & 43.37\small±1.13 & 43.70\small±1.10 \\
& \multicolumn{1}{|l}{} & \multicolumn{1}{|l|}{\textbf{Ours}} & \textbf{44.87\small±1.10} & \textbf{44.24\small±0.98}  \\

\bottomrule
\end{tabular}}
\label{tab:Efactresults}
\end{table}

%% file: tables/metadata.tex
\begin{table}
\centering
\caption{Effect of metadata: F1 results on the test set.}
\vspace{-4mm}
\scalebox{0.92}{
\begin{tabular}{l|cc}
\toprule
Extractors & mBERT-Based  & Graph-Based\\
\midrule
Snippets w/o Meta & 40.88 & 41.21\\
Snippets  & 42.14  & 43.53 \\
\midrule
Documents w/o Meta & 46.04 & 46.36 \\
Documents ($K$=30) & \textbf{46.42} & \textbf{46.89}\\
\midrule
Documents ($K$=10) & 43.85 & 45.83 \\
Documents ($K$=20) & 45.74  & 46.02 \\
Documents ($K$=50) & 45.59  & 46.35\\
\bottomrule
\end{tabular}}
\label{tab:meta}
\end{table}

%% file: tables/evidence_number.tex
\begin{table}
\centering
\caption{Effects of evidence: F1 results on the test set are reported. \#E indicates the number of evidence.}
\vspace{-4mm}
\scalebox{0.95}{
\begin{tabular}{lccccc}
\toprule
\textbf{\#E} & 1 & 3 & 5 & 10 & 15  \\
\midrule
Documents (Surface) & 41.06 & \textbf{42.84} & 42.76 & 42.26 & 41.64  \\
Documents (Semantic) & 41.63 & 42.35 & \textbf{42.89} & 42.40  & 42.13  \\
Documents (Hybrid) & 42.04 & 42.56 & \textbf{42.98} & 42.52  & 42.19   \\

\bottomrule
\end{tabular}}
\label{tab:evidence}
\vspace{-4mm}
\end{table}

%% file: tables/human.tex
\begin{table}
\centering
\caption{Human evaluation of extracted evidence. (5), (10), (15) denote the number of extracted sentences.}
\vspace{-4mm}
\scalebox{0.9}{
\begin{tabular}{l|ccc}
\toprule
Extractors & Precision & Recall & F1 \\
\midrule
Semantic (5) & 21.48 & 10.78 & 14.36 \\
Surface (5) & \textbf{26.68}  & 7.89  & 12.18 \\
Hybrid (5) & 23.35  & 9.98  & 13.98 \\
\midrule
Semantic (10) & 20.77 & 23.25 & 21.94 \\
Surface (10) & 19.50  & 28.55  & 23.17 \\
Hybrid (10) & 23.98  & 20.99  & 22.39 \\ 
\midrule
Semantic (15) & 11.27 & 42.64 & 17.83\\
Surface (15) & 9.05  & \textbf{48.77}  & 15.27 \\
Hybrid (15) & 12.47  & 52.66  & 20.16 \\
\midrule
HardKuma & 20.19  & 26.54  & 22.94\\
\textbf{Ours} & 22.79  & 33.76  & \textbf{27.21}\\
\bottomrule
\end{tabular}}

\vspace{-4mm}
\label{tab:ex}
\end{table}

%% file: subfiles/8_limitation.tex
\section{Limitations}
\label{limitation}

In this paper, we propose to incorporate full text of web pages for verifying real-world claims. Though the proposed fact-checking system significantly outperforms baselines, it still has the following three major limitations. Firstly, the training corpus only contains claims selected and verified by fact-checkers, as it is crawled from fact-checking agencies. Fact-checkers select and verify claims based on their judgements as well as public interests. Thus, there is no guarantee that the training corpus can cover any topics. Secondly, evidence in the retrieved web pages can be exhibited in the tables, PDFs, images, audios and videos. Human fact-checkers are able to extract relevant information from these heterogeneous sources, while our fact-checking system can only extract textual sentences as evidence. Unlike an artificial fact-checking dataset that assumes the world knowledge is restricted to Wikipedia, real-world dataset requires knowledge from more diversified sources. Using a search engine is an effective approach to obtain related knowledge, but it also brings the concern of untrustworthy evidence. Not all web documents returned by the search engine are equally trustworthy, and sometimes trustworthy sources contradict each other. Almost all existing fact-checking systems including ours are not able to address the presence of disagreeing or untrustworthy evidence. 

%% file: subfiles/6_conclusion.tex
\section{Conclusions}
\label{conclusion}

In this paper, we first analyzed the real-world dataset XFact, then proposed to incorporate retrieved documents as evidence to enrich the dataset. A latent variable model is further developed to jointly select evidence and predict factuality. Experiments indicate that retrieved documents can provide sufficient contextual clues to the model even when gold evidence sentences are not annotated. Our model maintains a balance between keeping relevant information and removing irrelevant information from source documents.



%% file: sample-authordraft.bbl

\begin{thebibliography}{56}


\ifx \showCODEN    \undefined \def \showCODEN     #1{\unskip}     \fi
\ifx \showDOI      \undefined \def \showDOI       #1{#1}\fi
\ifx \showISBNx    \undefined \def \showISBNx     #1{\unskip}     \fi
\ifx \showISBNxiii \undefined \def \showISBNxiii  #1{\unskip}     \fi
\ifx \showISSN     \undefined \def \showISSN      #1{\unskip}     \fi
\ifx \showLCCN     \undefined \def \showLCCN      #1{\unskip}     \fi
\ifx \shownote     \undefined \def \shownote      #1{#1}          \fi
\ifx \showarticletitle \undefined \def \showarticletitle #1{#1}   \fi
\ifx \showURL      \undefined \def \showURL       {\relax}        \fi
\providecommand\bibfield[2]{#2}
\providecommand\bibinfo[2]{#2}
\providecommand\natexlab[1]{#1}
\providecommand\showeprint[2][]{arXiv:#2}

\bibitem[Adair et~al\mbox{.}(2017)]%
        {Adair2017ProgressT}
\bibfield{author}{\bibinfo{person}{Bill Adair}, \bibinfo{person}{Chengkai Li}, \bibinfo{person}{Jun Yang}, {and} \bibinfo{person}{Cong Yu}.} \bibinfo{year}{2017}\natexlab{}.
\newblock \showarticletitle{Progress toward “the holy grail”: The continued quest to automate fact-checking}. In \bibinfo{booktitle}{\emph{Proceedings of the 2017 Computation+Journalism Symposium}}.
\newblock


\bibitem[Aly et~al\mbox{.}(2021)]%
        {Aly2021FEVEROUSFE}
\bibfield{author}{\bibinfo{person}{Rami Aly}, \bibinfo{person}{Zhijiang Guo}, \bibinfo{person}{M. Schlichtkrull}, \bibinfo{person}{James Thorne}, \bibinfo{person}{Andreas Vlachos}, \bibinfo{person}{Christos Christodoulopoulos}, \bibinfo{person}{O. Cocarascu}, {and} \bibinfo{person}{Arpit Mittal}.} \bibinfo{year}{2021}\natexlab{}.
\newblock \showarticletitle{{FEVEROUS}: {Fact Extraction} and {VERification} Over Unstructured and Structured information}.
\newblock \bibinfo{journal}{\emph{35th Conference on Neural Information Processing Systems (NeurIPS 2021) Track on Datasets and Benchmarks}} (\bibinfo{year}{2021}).
\newblock


\bibitem[Augenstein et~al\mbox{.}(2019)]%
        {Augenstein2019MultiFCAR}
\bibfield{author}{\bibinfo{person}{Isabelle Augenstein}, \bibinfo{person}{Christina Lioma}, \bibinfo{person}{Dongsheng Wang}, \bibinfo{person}{Lucas Chaves~Lima}, \bibinfo{person}{Casper Hansen}, \bibinfo{person}{Christian Hansen}, {and} \bibinfo{person}{Jakob~Grue Simonsen}.} \bibinfo{year}{2019}\natexlab{}.
\newblock \showarticletitle{{M}ulti{FC}: A Real-World Multi-Domain Dataset for Evidence-Based Fact Checking of Claims}. In \bibinfo{booktitle}{\emph{Proceedings of the 2019 Conference on Empirical Methods in Natural Language Processing and the 9th International Joint Conference on Natural Language Processing (EMNLP-IJCNLP)}}. \bibinfo{publisher}{Association for Computational Linguistics}, \bibinfo{address}{Hong Kong, China}, \bibinfo{pages}{4685--4697}.
\newblock
\urldef\tempurl%
\url{https://doi.org/10.18653/v1/D19-1475}
\showDOI{\tempurl}


\bibitem[Bastings et~al\mbox{.}(2019)]%
        {BastingsAT19}
\bibfield{author}{\bibinfo{person}{Jasmijn Bastings}, \bibinfo{person}{Wilker Aziz}, {and} \bibinfo{person}{Ivan Titov}.} \bibinfo{year}{2019}\natexlab{}.
\newblock \showarticletitle{Interpretable Neural Predictions with Differentiable Binary Variables}. In \bibinfo{booktitle}{\emph{Proceedings of the 57th Conference of the Association for Computational Linguistics, {ACL} 2019, Florence, Italy, July 28- August 2, 2019, Volume 1: Long Papers}}, \bibfield{editor}{\bibinfo{person}{Anna Korhonen}, \bibinfo{person}{David~R. Traum}, {and} \bibinfo{person}{Llu{\'{\i}}s M{\`{a}}rquez}} (Eds.). \bibinfo{publisher}{Association for Computational Linguistics}, \bibinfo{pages}{2963--2977}.
\newblock
\urldef\tempurl%
\url{https://doi.org/10.18653/v1/p19-1284}
\showDOI{\tempurl}


\bibitem[Chan et~al\mbox{.}(2022)]%
        {chan2022unirex}
\bibfield{author}{\bibinfo{person}{Aaron Chan}, \bibinfo{person}{Maziar Sanjabi}, \bibinfo{person}{Lambert Mathias}, \bibinfo{person}{Liang Tan}, \bibinfo{person}{Shaoliang Nie}, \bibinfo{person}{Xiaochang Peng}, \bibinfo{person}{Xiang Ren}, {and} \bibinfo{person}{Hamed Firooz}.} \bibinfo{year}{2022}\natexlab{}.
\newblock \showarticletitle{Unirex: A unified learning framework for language model rationale extraction}. In \bibinfo{booktitle}{\emph{International Conference on Machine Learning}}. PMLR, \bibinfo{pages}{2867--2889}.
\newblock


\bibitem[Chen et~al\mbox{.}(2022)]%
        {ChenSCD22}
\bibfield{author}{\bibinfo{person}{Jifan Chen}, \bibinfo{person}{Aniruddh Sriram}, \bibinfo{person}{Eunsol Choi}, {and} \bibinfo{person}{Greg Durrett}.} \bibinfo{year}{2022}\natexlab{}.
\newblock \showarticletitle{Generating Literal and Implied Subquestions to Fact-check Complex Claims}. In \bibinfo{booktitle}{\emph{Proceedings of the 2022 Conference on Empirical Methods in Natural Language Processing, {EMNLP} 2022, Abu Dhabi, United Arab Emirates, December 7-11, 2022}}, \bibfield{editor}{\bibinfo{person}{Yoav Goldberg}, \bibinfo{person}{Zornitsa Kozareva}, {and} \bibinfo{person}{Yue Zhang}} (Eds.). \bibinfo{publisher}{Association for Computational Linguistics}, \bibinfo{pages}{3495--3516}.
\newblock
\urldef\tempurl%
\url{https://doi.org/10.18653/v1/2022.emnlp-main.229}
\showDOI{\tempurl}


\bibitem[Chen et~al\mbox{.}(2020)]%
        {Chen2020TabFactAL}
\bibfield{author}{\bibinfo{person}{Wenhu Chen}, \bibinfo{person}{Hongmin Wang}, \bibinfo{person}{Jianshu Chen}, \bibinfo{person}{Yunkai Zhang}, \bibinfo{person}{Hong Wang}, \bibinfo{person}{Shiyang Li}, \bibinfo{person}{Xiyou Zhou}, {and} \bibinfo{person}{William~Yang Wang}.} \bibinfo{year}{2020}\natexlab{}.
\newblock \showarticletitle{{TabFact}: {A} Large-scale Dataset for Table-based Fact Verification}. In \bibinfo{booktitle}{\emph{8th International Conference on Learning Representations, {ICLR} 2020}}. \bibinfo{address}{Addis Ababa, Ethiopia}.
\newblock
\urldef\tempurl%
\url{https://openreview.net/forum?id=rkeJRhNYDH}
\showURL{%
\tempurl}


\bibitem[Devlin et~al\mbox{.}(2019)]%
        {Devlin2019BERTPO}
\bibfield{author}{\bibinfo{person}{Jacob Devlin}, \bibinfo{person}{Ming-Wei Chang}, \bibinfo{person}{Kenton Lee}, {and} \bibinfo{person}{Kristina Toutanova}.} \bibinfo{year}{2019}\natexlab{}.
\newblock \showarticletitle{{BERT}: Pre-training of Deep Bidirectional Transformers for Language Understanding}. In \bibinfo{booktitle}{\emph{Proceedings of the 2019 Conference of the North {A}merican Chapter of the Association for Computational Linguistics: Human Language Technologies, Volume 1 (Long and Short Papers)}}. \bibinfo{publisher}{Association for Computational Linguistics}, \bibinfo{address}{Minneapolis, Minnesota}, \bibinfo{pages}{4171--4186}.
\newblock
\urldef\tempurl%
\url{https://doi.org/10.18653/v1/N19-1423}
\showDOI{\tempurl}


\bibitem[Ferreira and Vlachos(2016)]%
        {Ferreira2016EmergentAN}
\bibfield{author}{\bibinfo{person}{William Ferreira} {and} \bibinfo{person}{Andreas Vlachos}.} \bibinfo{year}{2016}\natexlab{}.
\newblock \showarticletitle{{E}mergent: a novel data-set for stance classification}. In \bibinfo{booktitle}{\emph{Proceedings of the 2016 Conference of the North {A}merican Chapter of the Association for Computational Linguistics: Human Language Technologies}}. \bibinfo{publisher}{Association for Computational Linguistics}, \bibinfo{address}{San Diego, California}, \bibinfo{pages}{1163--1168}.
\newblock
\urldef\tempurl%
\url{https://doi.org/10.18653/v1/N16-1138}
\showDOI{\tempurl}


\bibitem[Fleiss(1971)]%
        {fleiss1971measuring}
\bibfield{author}{\bibinfo{person}{Joseph~L Fleiss}.} \bibinfo{year}{1971}\natexlab{}.
\newblock \showarticletitle{Measuring nominal scale agreement among many raters.}
\newblock \bibinfo{journal}{\emph{Psychological bulletin}} \bibinfo{volume}{76}, \bibinfo{number}{5} (\bibinfo{year}{1971}), \bibinfo{pages}{378}.
\newblock


\bibitem[Graves(2018)]%
        {2018graves}
\bibfield{author}{\bibinfo{person}{Lucas Graves}.} \bibinfo{year}{2018}\natexlab{}.
\newblock \showarticletitle{Understanding the Promise and Limits of Automated Fact-checking}.
\newblock \bibinfo{journal}{\emph{Reuters Institute for the Study of Journalism}} (\bibinfo{year}{2018}).
\newblock


\bibitem[Guo et~al\mbox{.}(2021)]%
        {guo2021}
\bibfield{author}{\bibinfo{person}{Zhijiang Guo}, \bibinfo{person}{Michael~Sejr Schlichtkrull}, {and} \bibinfo{person}{Andreas Vlachos}.} \bibinfo{year}{2021}\natexlab{}.
\newblock \showarticletitle{A Survey on Automated Fact-Checking}.
\newblock \bibinfo{journal}{\emph{Transactions of the Association for Computational Linguistics}} (\bibinfo{year}{2021}).
\newblock


\bibitem[Gupta and Srikumar(2021)]%
        {gupta-srikumar-2021-x}
\bibfield{author}{\bibinfo{person}{Ashim Gupta} {and} \bibinfo{person}{Vivek Srikumar}.} \bibinfo{year}{2021}\natexlab{}.
\newblock \showarticletitle{{X-Fact}: A New Benchmark Dataset for Multilingual Fact Checking}. In \bibinfo{booktitle}{\emph{Proceedings of the 59th Annual Meeting of the Association for Computational Linguistics and the 11th International Joint Conference on Natural Language Processing (Volume 2: Short Papers)}}. \bibinfo{publisher}{Association for Computational Linguistics}, \bibinfo{address}{Online}, \bibinfo{pages}{675--682}.
\newblock
\urldef\tempurl%
\url{https://doi.org/10.18653/v1/2021.acl-short.86}
\showDOI{\tempurl}


\bibitem[Gupta et~al\mbox{.}(2020)]%
        {Gupta2020INFOTABSIO}
\bibfield{author}{\bibinfo{person}{Vivek Gupta}, \bibinfo{person}{Maitrey Mehta}, \bibinfo{person}{Pegah Nokhiz}, {and} \bibinfo{person}{Vivek Srikumar}.} \bibinfo{year}{2020}\natexlab{}.
\newblock \showarticletitle{{INFOTABS}: Inference on Tables as Semi-structured Data}. In \bibinfo{booktitle}{\emph{Proceedings of the 58th Annual Meeting of the Association for Computational Linguistics}}. \bibinfo{publisher}{Association for Computational Linguistics}, \bibinfo{address}{Online}, \bibinfo{pages}{2309--2324}.
\newblock
\urldef\tempurl%
\url{https://doi.org/10.18653/v1/2020.acl-main.210}
\showDOI{\tempurl}


\bibitem[Hanselowski et~al\mbox{.}(2019)]%
        {Hanselowski2019ARA}
\bibfield{author}{\bibinfo{person}{Andreas Hanselowski}, \bibinfo{person}{Christian Stab}, \bibinfo{person}{Claudia Schulz}, \bibinfo{person}{Zile Li}, {and} \bibinfo{person}{Iryna Gurevych}.} \bibinfo{year}{2019}\natexlab{}.
\newblock \showarticletitle{A Richly Annotated Corpus for Different Tasks in Automated Fact-Checking}. In \bibinfo{booktitle}{\emph{Proceedings of the 23rd Conference on Computational Natural Language Learning (CoNLL)}}. \bibinfo{publisher}{Association for Computational Linguistics}, \bibinfo{address}{Hong Kong, China}, \bibinfo{pages}{493--503}.
\newblock
\urldef\tempurl%
\url{https://doi.org/10.18653/v1/K19-1046}
\showDOI{\tempurl}


\bibitem[Hanselowski et~al\mbox{.}(2018)]%
        {hanselowski-etal-2018-ukp}
\bibfield{author}{\bibinfo{person}{Andreas Hanselowski}, \bibinfo{person}{Hao Zhang}, \bibinfo{person}{Zile Li}, \bibinfo{person}{Daniil Sorokin}, \bibinfo{person}{Benjamin Schiller}, \bibinfo{person}{Claudia Schulz}, {and} \bibinfo{person}{Iryna Gurevych}.} \bibinfo{year}{2018}\natexlab{}.
\newblock \showarticletitle{{UKP}-Athene: Multi-Sentence Textual Entailment for Claim Verification}. In \bibinfo{booktitle}{\emph{Proceedings of the First Workshop on Fact Extraction and {VER}ification ({FEVER})}}. \bibinfo{publisher}{Association for Computational Linguistics}, \bibinfo{address}{Brussels, Belgium}, \bibinfo{pages}{103--108}.
\newblock
\urldef\tempurl%
\url{https://doi.org/10.18653/v1/W18-5516}
\showDOI{\tempurl}


\bibitem[Huang et~al\mbox{.}(2022)]%
        {huang2022concrete}
\bibfield{author}{\bibinfo{person}{Kung-Hsiang Huang}, \bibinfo{person}{ChengXiang Zhai}, {and} \bibinfo{person}{Heng Ji}.} \bibinfo{year}{2022}\natexlab{}.
\newblock \showarticletitle{CONCRETE: Improving Cross-lingual Fact-checking with Cross-lingual Retrieval}. In \bibinfo{booktitle}{\emph{Proceedings of the 29th International Conference on Computational Linguistics}}. \bibinfo{pages}{1024--1035}.
\newblock


\bibitem[Jain et~al\mbox{.}(2020)]%
        {jain2020learning}
\bibfield{author}{\bibinfo{person}{Sarthak Jain}, \bibinfo{person}{Sarah Wiegreffe}, \bibinfo{person}{Yuval Pinter}, {and} \bibinfo{person}{Byron~C Wallace}.} \bibinfo{year}{2020}\natexlab{}.
\newblock \showarticletitle{Learning to Faithfully Rationalize by Construction}. In \bibinfo{booktitle}{\emph{Proceedings of the 58th Annual Meeting of the Association for Computational Linguistics}}. \bibinfo{pages}{4459--4473}.
\newblock


\bibitem[Jiang et~al\mbox{.}(2020)]%
        {jiang2020hover}
\bibfield{author}{\bibinfo{person}{Yichen Jiang}, \bibinfo{person}{Shikha Bordia}, \bibinfo{person}{Zheng Zhong}, \bibinfo{person}{Charles Dognin}, \bibinfo{person}{Maneesh Singh}, {and} \bibinfo{person}{Mohit Bansal}.} \bibinfo{year}{2020}\natexlab{}.
\newblock \showarticletitle{{H}o{V}er: A Dataset for Many-Hop Fact Extraction And Claim Verification}. In \bibinfo{booktitle}{\emph{Findings of the Association for Computational Linguistics: EMNLP 2020}}. \bibinfo{publisher}{Association for Computational Linguistics}, \bibinfo{address}{Online}, \bibinfo{pages}{3441--3460}.
\newblock
\urldef\tempurl%
\url{https://doi.org/10.18653/v1/2020.findings-emnlp.309}
\showDOI{\tempurl}


\bibitem[Khan et~al\mbox{.}(2022)]%
        {KhanWP22}
\bibfield{author}{\bibinfo{person}{Kashif Khan}, \bibinfo{person}{Ruizhe Wang}, {and} \bibinfo{person}{Pascal Poupart}.} \bibinfo{year}{2022}\natexlab{}.
\newblock \showarticletitle{WatClaimCheck: {A} new Dataset for Claim Entailment and Inference}. In \bibinfo{booktitle}{\emph{Proceedings of the 60th Annual Meeting of the Association for Computational Linguistics (Volume 1: Long Papers), {ACL} 2022, Dublin, Ireland, May 22-27, 2022}}, \bibfield{editor}{\bibinfo{person}{Smaranda Muresan}, \bibinfo{person}{Preslav Nakov}, {and} \bibinfo{person}{Aline Villavicencio}} (Eds.). \bibinfo{publisher}{Association for Computational Linguistics}, \bibinfo{pages}{1293--1304}.
\newblock
\urldef\tempurl%
\url{https://doi.org/10.18653/v1/2022.acl-long.92}
\showDOI{\tempurl}


\bibitem[Khouja(2020)]%
        {Khouja2020StancePA}
\bibfield{author}{\bibinfo{person}{Jude Khouja}.} \bibinfo{year}{2020}\natexlab{}.
\newblock \showarticletitle{Stance Prediction and Claim Verification: An {A}rabic Perspective}. In \bibinfo{booktitle}{\emph{Proceedings of the Third Workshop on Fact Extraction and VERification (FEVER)}}. \bibinfo{publisher}{Association for Computational Linguistics}, \bibinfo{address}{Online}, \bibinfo{pages}{8--17}.
\newblock
\urldef\tempurl%
\url{https://doi.org/10.18653/v1/2020.fever-1.2}
\showDOI{\tempurl}


\bibitem[Kingma and Ba(2015)]%
        {kingma2014adam}
\bibfield{author}{\bibinfo{person}{Diederik~P Kingma} {and} \bibinfo{person}{Jimmy Ba}.} \bibinfo{year}{2015}\natexlab{}.
\newblock \showarticletitle{Adam: A method for stochastic optimization}. In \bibinfo{booktitle}{\emph{Proc. of ICLR}}.
\newblock


\bibitem[Kingma and Welling(2014)]%
        {KingmaW14}
\bibfield{author}{\bibinfo{person}{Diederik~P. Kingma} {and} \bibinfo{person}{Max Welling}.} \bibinfo{year}{2014}\natexlab{}.
\newblock \showarticletitle{Auto-Encoding Variational Bayes}. In \bibinfo{booktitle}{\emph{2nd International Conference on Learning Representations, {ICLR} 2014, Banff, AB, Canada, April 14-16, 2014, Conference Track Proceedings}}, \bibfield{editor}{\bibinfo{person}{Yoshua Bengio} {and} \bibinfo{person}{Yann LeCun}} (Eds.).
\newblock
\urldef\tempurl%
\url{http://arxiv.org/abs/1312.6114}
\showURL{%
\tempurl}


\bibitem[Koo et~al\mbox{.}(2010)]%
        {KooRCJS10}
\bibfield{author}{\bibinfo{person}{Terry Koo}, \bibinfo{person}{Alexander~M. Rush}, \bibinfo{person}{Michael Collins}, \bibinfo{person}{Tommi~S. Jaakkola}, {and} \bibinfo{person}{David~A. Sontag}.} \bibinfo{year}{2010}\natexlab{}.
\newblock \showarticletitle{Dual Decomposition for Parsing with Non-Projective Head Automata}. In \bibinfo{booktitle}{\emph{Proceedings of the 2010 Conference on Empirical Methods in Natural Language Processing, {EMNLP} 2010, 9-11 October 2010, {MIT} Stata Center, Massachusetts, USA, {A} meeting of SIGDAT, a Special Interest Group of the {ACL}}}. \bibinfo{publisher}{{ACL}}, \bibinfo{pages}{1288--1298}.
\newblock
\urldef\tempurl%
\url{https://aclanthology.org/D10-1125/}
\showURL{%
\tempurl}


\bibitem[Kotonya and Toni(2020)]%
        {kotonya2020health}
\bibfield{author}{\bibinfo{person}{Neema Kotonya} {and} \bibinfo{person}{Francesca Toni}.} \bibinfo{year}{2020}\natexlab{}.
\newblock \showarticletitle{Explainable Automated Fact-Checking for Public Health Claims}. In \bibinfo{booktitle}{\emph{Proceedings of the 2020 Conference on Empirical Methods in Natural Language Processing (EMNLP)}}. \bibinfo{publisher}{Association for Computational Linguistics}, \bibinfo{address}{Online}, \bibinfo{pages}{7740--7754}.
\newblock
\urldef\tempurl%
\url{https://doi.org/10.18653/v1/2020.emnlp-main.623}
\showDOI{\tempurl}


\bibitem[Lei et~al\mbox{.}(2016)]%
        {LeiBJ16}
\bibfield{author}{\bibinfo{person}{Tao Lei}, \bibinfo{person}{Regina Barzilay}, {and} \bibinfo{person}{Tommi~S. Jaakkola}.} \bibinfo{year}{2016}\natexlab{}.
\newblock \showarticletitle{Rationalizing Neural Predictions}. In \bibinfo{booktitle}{\emph{Proceedings of the 2016 Conference on Empirical Methods in Natural Language Processing, {EMNLP} 2016, Austin, Texas, USA, November 1-4, 2016}}, \bibfield{editor}{\bibinfo{person}{Jian Su}, \bibinfo{person}{Xavier Carreras}, {and} \bibinfo{person}{Kevin Duh}} (Eds.). \bibinfo{publisher}{The Association for Computational Linguistics}, \bibinfo{pages}{107--117}.
\newblock
\urldef\tempurl%
\url{https://doi.org/10.18653/v1/d16-1011}
\showDOI{\tempurl}


\bibitem[Liu et~al\mbox{.}(2024)]%
        {liu2024enhancing}
\bibfield{author}{\bibinfo{person}{Wei Liu}, \bibinfo{person}{Haozhao Wang}, \bibinfo{person}{Jun Wang}, \bibinfo{person}{Zhiying Deng}, \bibinfo{person}{YuanKai Zhang}, \bibinfo{person}{Cheng Wang}, {and} \bibinfo{person}{Ruixuan Li}.} \bibinfo{year}{2024}\natexlab{}.
\newblock \showarticletitle{Enhancing the Rationale-Input Alignment for Self-explaining Rationalization}. In \bibinfo{booktitle}{\emph{Proceedings of the 40th IEEE International Conference on Data Engineering (ICDE)}}.
\newblock


\bibitem[Liu et~al\mbox{.}(2020)]%
        {Liu2020KernelGA}
\bibfield{author}{\bibinfo{person}{Zhenghao Liu}, \bibinfo{person}{Chenyan Xiong}, \bibinfo{person}{Maosong Sun}, {and} \bibinfo{person}{Zhiyuan Liu}.} \bibinfo{year}{2020}\natexlab{}.
\newblock \showarticletitle{Fine-grained Fact Verification with Kernel Graph Attention Network}. In \bibinfo{booktitle}{\emph{Proceedings of the 58th Annual Meeting of the Association for Computational Linguistics}}. \bibinfo{publisher}{Association for Computational Linguistics}, \bibinfo{address}{Online}, \bibinfo{pages}{7342--7351}.
\newblock
\urldef\tempurl%
\url{https://doi.org/10.18653/v1/2020.acl-main.655}
\showDOI{\tempurl}


\bibitem[Luken et~al\mbox{.}(2018)]%
        {luken-etal-2018-qed}
\bibfield{author}{\bibinfo{person}{Jackson Luken}, \bibinfo{person}{Nanjiang Jiang}, {and} \bibinfo{person}{Marie-Catherine de Marneffe}.} \bibinfo{year}{2018}\natexlab{}.
\newblock \showarticletitle{{QED}: A fact verification system for the {FEVER} shared task}. In \bibinfo{booktitle}{\emph{Proceedings of the First Workshop on Fact Extraction and {VER}ification ({FEVER})}}. \bibinfo{publisher}{Association for Computational Linguistics}, \bibinfo{address}{Brussels, Belgium}, \bibinfo{pages}{156--160}.
\newblock
\urldef\tempurl%
\url{https://doi.org/10.18653/v1/W18-5526}
\showDOI{\tempurl}


\bibitem[Maddison et~al\mbox{.}(2017)]%
        {maddisonMT17}
\bibfield{author}{\bibinfo{person}{Chris~J. Maddison}, \bibinfo{person}{Andriy Mnih}, {and} \bibinfo{person}{Yee~Whye Teh}.} \bibinfo{year}{2017}\natexlab{}.
\newblock \showarticletitle{The Concrete Distribution: {A} Continuous Relaxation of Discrete Random Variables}. In \bibinfo{booktitle}{\emph{5th International Conference on Learning Representations, {ICLR} 2017, Toulon, France, April 24-26, 2017, Conference Track Proceedings}}. \bibinfo{publisher}{OpenReview.net}.
\newblock
\urldef\tempurl%
\url{https://openreview.net/forum?id=S1jE5L5gl}
\showURL{%
\tempurl}


\bibitem[Martins et~al\mbox{.}(2015)]%
        {MartinsFASX15}
\bibfield{author}{\bibinfo{person}{Andr{\'{e}} F.~T. Martins}, \bibinfo{person}{M{\'{a}}rio A.~T. Figueiredo}, \bibinfo{person}{Pedro M.~Q. Aguiar}, \bibinfo{person}{Noah~A. Smith}, {and} \bibinfo{person}{Eric~P. Xing}.} \bibinfo{year}{2015}\natexlab{}.
\newblock \showarticletitle{AD\({}^{\mbox{3}}\): alternating directions dual decomposition for {MAP} inference in graphical models}.
\newblock \bibinfo{journal}{\emph{J. Mach. Learn. Res.}}  \bibinfo{volume}{16} (\bibinfo{year}{2015}), \bibinfo{pages}{495--545}.
\newblock
\urldef\tempurl%
\url{http://dl.acm.org/citation.cfm?id=2789288}
\showURL{%
\tempurl}


\bibitem[Niculae and Blondel(2017)]%
        {NiculaeB17}
\bibfield{author}{\bibinfo{person}{Vlad Niculae} {and} \bibinfo{person}{Mathieu Blondel}.} \bibinfo{year}{2017}\natexlab{}.
\newblock \showarticletitle{A Regularized Framework for Sparse and Structured Neural Attention}. In \bibinfo{booktitle}{\emph{Advances in Neural Information Processing Systems 30: Annual Conference on Neural Information Processing Systems 2017, December 4-9, 2017, Long Beach, CA, {USA}}}, \bibfield{editor}{\bibinfo{person}{Isabelle Guyon}, \bibinfo{person}{Ulrike von Luxburg}, \bibinfo{person}{Samy Bengio}, \bibinfo{person}{Hanna~M. Wallach}, \bibinfo{person}{Rob Fergus}, \bibinfo{person}{S.~V.~N. Vishwanathan}, {and} \bibinfo{person}{Roman Garnett}} (Eds.). \bibinfo{pages}{3338--3348}.
\newblock
\urldef\tempurl%
\url{https://proceedings.neurips.cc/paper/2017/hash/2d1b2a5ff364606ff041650887723470-Abstract.html}
\showURL{%
\tempurl}


\bibitem[Niculae and Martins(2020)]%
        {NiculaeM20}
\bibfield{author}{\bibinfo{person}{Vlad Niculae} {and} \bibinfo{person}{Andr{\'{e}} F.~T. Martins}.} \bibinfo{year}{2020}\natexlab{}.
\newblock \showarticletitle{LP-SparseMAP: Differentiable Relaxed Optimization for Sparse Structured Prediction}. In \bibinfo{booktitle}{\emph{Proceedings of the 37th International Conference on Machine Learning, {ICML} 2020, 13-18 July 2020, Virtual Event}} \emph{(\bibinfo{series}{Proceedings of Machine Learning Research}, Vol.~\bibinfo{volume}{119})}. \bibinfo{publisher}{{PMLR}}, \bibinfo{pages}{7348--7359}.
\newblock
\urldef\tempurl%
\url{http://proceedings.mlr.press/v119/niculae20a.html}
\showURL{%
\tempurl}


\bibitem[Nie et~al\mbox{.}(2019)]%
        {Nie2019CombiningFE}
\bibfield{author}{\bibinfo{person}{Yixin Nie}, \bibinfo{person}{Haonan Chen}, {and} \bibinfo{person}{Mohit Bansal}.} \bibinfo{year}{2019}\natexlab{}.
\newblock \showarticletitle{Combining Fact Extraction and Verification with Neural Semantic Matching Networks}. In \bibinfo{booktitle}{\emph{The Thirty-Third {AAAI} Conference on Artificial Intelligence, {AAAI} 2019, The Thirty-First Innovative Applications of Artificial Intelligence Conference, {IAAI} 2019, The Ninth {AAAI} Symposium on Educational Advances in Artificial Intelligence, {EAAI} 2019, Honolulu, Hawaii, USA, January 27 - February 1, 2019}}. \bibinfo{publisher}{{AAAI} Press}, \bibinfo{pages}{6859--6866}.
\newblock
\urldef\tempurl%
\url{https://doi.org/10.1609/aaai.v33i01.33016859}
\showDOI{\tempurl}


\bibitem[Pomerleau and Rao(2017)]%
        {pomerleau2017fake}
\bibfield{author}{\bibinfo{person}{Dean Pomerleau} {and} \bibinfo{person}{Delip Rao}.} \bibinfo{year}{2017}\natexlab{}.
\newblock \showarticletitle{The fake news challenge: Exploring how artificial intelligence technologies could be leveraged to combat fake news}.
\newblock \bibinfo{journal}{\emph{Fake News Challenge}} (\bibinfo{year}{2017}).
\newblock


\bibitem[Rashkin et~al\mbox{.}(2017)]%
        {Rashkin2017TruthOV}
\bibfield{author}{\bibinfo{person}{Hannah Rashkin}, \bibinfo{person}{Eunsol Choi}, \bibinfo{person}{Jin~Yea Jang}, \bibinfo{person}{Svitlana Volkova}, {and} \bibinfo{person}{Yejin Choi}.} \bibinfo{year}{2017}\natexlab{}.
\newblock \showarticletitle{Truth of Varying Shades: Analyzing Language in Fake News and Political Fact-Checking}. In \bibinfo{booktitle}{\emph{Proceedings of the 2017 Conference on Empirical Methods in Natural Language Processing}}. \bibinfo{publisher}{Association for Computational Linguistics}, \bibinfo{address}{Copenhagen, Denmark}, \bibinfo{pages}{2931--2937}.
\newblock
\urldef\tempurl%
\url{https://doi.org/10.18653/v1/D17-1317}
\showDOI{\tempurl}


\bibitem[Schlichtkrull et~al\mbox{.}(2023)]%
        {schlichtkrull2023averitec}
\bibfield{author}{\bibinfo{person}{Michael Schlichtkrull}, \bibinfo{person}{Zhijiang Guo}, {and} \bibinfo{person}{Andreas Vlachos}.} \bibinfo{year}{2023}\natexlab{}.
\newblock \showarticletitle{AVeriTeC: A dataset for real-world claim verification with evidence from the web}.
\newblock \bibinfo{journal}{\emph{arXiv preprint arXiv:2305.13117}} (\bibinfo{year}{2023}).
\newblock


\bibitem[Schlichtkrull et~al\mbox{.}(2021)]%
        {schlichtkrull2020}
\bibfield{author}{\bibinfo{person}{Michael~Sejr Schlichtkrull}, \bibinfo{person}{Vladimir Karpukhin}, \bibinfo{person}{Barlas Oguz}, \bibinfo{person}{Mike Lewis}, \bibinfo{person}{Wen-tau Yih}, {and} \bibinfo{person}{Sebastian Riedel}.} \bibinfo{year}{2021}\natexlab{}.
\newblock \showarticletitle{Joint Verification and Reranking for Open Fact Checking Over Tables}. In \bibinfo{booktitle}{\emph{Proceedings of the 59th Annual Meeting of the Association for Computational Linguistics and the 11th International Joint Conference on Natural Language Processing (Volume 1: Long Papers)}}. \bibinfo{publisher}{Association for Computational Linguistics}, \bibinfo{address}{Online}, \bibinfo{pages}{6787--6799}.
\newblock
\urldef\tempurl%
\url{https://doi.org/10.18653/v1/2021.acl-long.529}
\showDOI{\tempurl}


\bibitem[Schuster et~al\mbox{.}(2021)]%
        {vitaminc2021}
\bibfield{author}{\bibinfo{person}{Tal Schuster}, \bibinfo{person}{Adam Fisch}, {and} \bibinfo{person}{Regina Barzilay}.} \bibinfo{year}{2021}\natexlab{}.
\newblock \showarticletitle{Get Your {Vitamin C}! Robust Fact Verification with Contrastive Evidence}. In \bibinfo{booktitle}{\emph{Proceedings of the 2021 Conference of the North American Chapter of the Association for Computational Linguistics: Human Language Technologies}}. \bibinfo{publisher}{Association for Computational Linguistics}, \bibinfo{address}{Online}, \bibinfo{pages}{624--643}.
\newblock
\urldef\tempurl%
\url{https://www.aclweb.org/anthology/2021.naacl-main.52}
\showURL{%
\tempurl}


\bibitem[Schuster et~al\mbox{.}(2020)]%
        {Schuster2020TheLO}
\bibfield{author}{\bibinfo{person}{Tal Schuster}, \bibinfo{person}{Roei Schuster}, \bibinfo{person}{Darsh~J. Shah}, {and} \bibinfo{person}{Regina Barzilay}.} \bibinfo{year}{2020}\natexlab{}.
\newblock \showarticletitle{The Limitations of Stylometry for Detecting Machine-Generated Fake News}.
\newblock \bibinfo{journal}{\emph{Computational Linguistics}} \bibinfo{volume}{46}, \bibinfo{number}{2} (\bibinfo{year}{2020}), \bibinfo{pages}{499--510}.
\newblock
\urldef\tempurl%
\url{https://doi.org/10.1162/coli_a_00380}
\showDOI{\tempurl}


\bibitem[Shaar et~al\mbox{.}(2020)]%
        {Shaar2020ThatIA}
\bibfield{author}{\bibinfo{person}{Shaden Shaar}, \bibinfo{person}{Nikolay Babulkov}, \bibinfo{person}{Giovanni Da~San~Martino}, {and} \bibinfo{person}{Preslav Nakov}.} \bibinfo{year}{2020}\natexlab{}.
\newblock \showarticletitle{That is a Known Lie: Detecting Previously Fact-Checked Claims}. In \bibinfo{booktitle}{\emph{Proceedings of the 58th Annual Meeting of the Association for Computational Linguistics}}. \bibinfo{publisher}{Association for Computational Linguistics}, \bibinfo{address}{Online}, \bibinfo{pages}{3607--3618}.
\newblock
\urldef\tempurl%
\url{https://doi.org/10.18653/v1/2020.acl-main.332}
\showDOI{\tempurl}


\bibitem[Shahi and Nandini(2020)]%
        {shahifakecovid}
\bibfield{author}{\bibinfo{person}{Gautam~Kishore Shahi} {and} \bibinfo{person}{Durgesh Nandini}.} \bibinfo{year}{2020}\natexlab{}.
\newblock \showarticletitle{Fake{C}ovid -- A Multilingual Cross-domain Fact Check News Dataset for COVID-19}. In \bibinfo{booktitle}{\emph{Workshop Proceedings of the 14th International {AAAI} {C}onference on {W}eb and {S}ocial {M}edia}}.
\newblock
\urldef\tempurl%
\url{http://workshop-proceedings.icwsm.org/pdf/2020_14.pdf}
\showURL{%
\tempurl}


\bibitem[Thorne et~al\mbox{.}(2018a)]%
        {Thorne2018FEVERAL}
\bibfield{author}{\bibinfo{person}{James Thorne}, \bibinfo{person}{Andreas Vlachos}, \bibinfo{person}{Christos Christodoulopoulos}, {and} \bibinfo{person}{Arpit Mittal}.} \bibinfo{year}{2018}\natexlab{a}.
\newblock \showarticletitle{{FEVER}: a Large-scale Dataset for Fact Extraction and {VER}ification}. In \bibinfo{booktitle}{\emph{Proceedings of the 2018 Conference of the North {A}merican Chapter of the Association for Computational Linguistics: Human Language Technologies, Volume 1 (Long Papers)}}. \bibinfo{publisher}{Association for Computational Linguistics}, \bibinfo{address}{New Orleans, Louisiana}, \bibinfo{pages}{809--819}.
\newblock
\urldef\tempurl%
\url{https://doi.org/10.18653/v1/N18-1074}
\showDOI{\tempurl}


\bibitem[Thorne et~al\mbox{.}(2018b)]%
        {Thorne18Fact}
\bibfield{author}{\bibinfo{person}{James Thorne}, \bibinfo{person}{Andreas Vlachos}, \bibinfo{person}{Oana Cocarascu}, \bibinfo{person}{Christos Christodoulopoulos}, {and} \bibinfo{person}{Arpit Mittal}.} \bibinfo{year}{2018}\natexlab{b}.
\newblock \showarticletitle{The Fact Extraction and {VER}ification ({FEVER}) Shared Task}. In \bibinfo{booktitle}{\emph{Proceedings of the First Workshop on Fact Extraction and {VER}ification ({FEVER})}}. \bibinfo{publisher}{Association for Computational Linguistics}, \bibinfo{address}{Brussels, Belgium}, \bibinfo{pages}{1--9}.
\newblock
\urldef\tempurl%
\url{https://doi.org/10.18653/v1/W18-5501}
\showDOI{\tempurl}


\bibitem[Turenne(2018)]%
        {turenne2018rumour}
\bibfield{author}{\bibinfo{person}{Nicolas Turenne}.} \bibinfo{year}{2018}\natexlab{}.
\newblock \showarticletitle{The rumour spectrum}.
\newblock \bibinfo{journal}{\emph{PloS one}} \bibinfo{volume}{13}, \bibinfo{number}{1} (\bibinfo{year}{2018}), \bibinfo{pages}{e0189080}.
\newblock


\bibitem[Vosoughi et~al\mbox{.}(2018)]%
        {vosoughi2018spread}
\bibfield{author}{\bibinfo{person}{Soroush Vosoughi}, \bibinfo{person}{Deb Roy}, {and} \bibinfo{person}{Sinan Aral}.} \bibinfo{year}{2018}\natexlab{}.
\newblock \showarticletitle{The spread of true and false news online}.
\newblock \bibinfo{journal}{\emph{Science}} \bibinfo{volume}{359}, \bibinfo{number}{6380} (\bibinfo{year}{2018}), \bibinfo{pages}{1146--1151}.
\newblock


\bibitem[Wadden et~al\mbox{.}(2020)]%
        {Wadden2020FactOF}
\bibfield{author}{\bibinfo{person}{David Wadden}, \bibinfo{person}{Shanchuan Lin}, \bibinfo{person}{Kyle Lo}, \bibinfo{person}{Lucy~Lu Wang}, \bibinfo{person}{Madeleine van Zuylen}, \bibinfo{person}{Arman Cohan}, {and} \bibinfo{person}{Hannaneh Hajishirzi}.} \bibinfo{year}{2020}\natexlab{}.
\newblock \showarticletitle{{Fact or Fiction}: Verifying Scientific Claims}. In \bibinfo{booktitle}{\emph{Proceedings of the 2020 Conference on Empirical Methods in Natural Language Processing (EMNLP)}}. \bibinfo{publisher}{Association for Computational Linguistics}, \bibinfo{address}{Online}, \bibinfo{pages}{7534--7550}.
\newblock
\urldef\tempurl%
\url{https://doi.org/10.18653/v1/2020.emnlp-main.609}
\showDOI{\tempurl}


\bibitem[Wainwright and Jordan(2008)]%
        {Wainwright2008GraphicalME}
\bibfield{author}{\bibinfo{person}{Martin~J. Wainwright} {and} \bibinfo{person}{M.I. Jordan}.} \bibinfo{year}{2008}\natexlab{}.
\newblock \showarticletitle{Graphical Models, Exponential Families, and Variational Inference}.
\newblock \bibinfo{journal}{\emph{Found. Trends Mach. Learn.}}  \bibinfo{volume}{1} (\bibinfo{year}{2008}), \bibinfo{pages}{1--305}.
\newblock


\bibitem[Wang(2017)]%
        {Wang2017LiarLP}
\bibfield{author}{\bibinfo{person}{William~Yang Wang}.} \bibinfo{year}{2017}\natexlab{}.
\newblock \showarticletitle{{``}{Liar, Liar Pants on Fire{''}}: A New Benchmark Dataset for Fake News Detection}. In \bibinfo{booktitle}{\emph{Proceedings of the 55th Annual Meeting of the Association for Computational Linguistics (Volume 2: Short Papers)}}. \bibinfo{publisher}{Association for Computational Linguistics}, \bibinfo{address}{Vancouver, Canada}, \bibinfo{pages}{422--426}.
\newblock
\urldef\tempurl%
\url{https://doi.org/10.18653/v1/P17-2067}
\showDOI{\tempurl}


\bibitem[Williams(1992)]%
        {williams1992simple}
\bibfield{author}{\bibinfo{person}{Ronald~J Williams}.} \bibinfo{year}{1992}\natexlab{}.
\newblock \showarticletitle{Simple statistical gradient-following algorithms for connectionist reinforcement learning}.
\newblock \bibinfo{journal}{\emph{Machine learning}} \bibinfo{volume}{8}, \bibinfo{number}{3} (\bibinfo{year}{1992}), \bibinfo{pages}{229--256}.
\newblock


\bibitem[Wu et~al\mbox{.}(2023)]%
        {wu2023adversarial}
\bibfield{author}{\bibinfo{person}{Junfei Wu}, \bibinfo{person}{Weizhi Xu}, \bibinfo{person}{Qiang Liu}, \bibinfo{person}{Shu Wu}, {and} \bibinfo{person}{Liang Wang}.} \bibinfo{year}{2023}\natexlab{}.
\newblock \showarticletitle{Adversarial contrastive learning for evidence-aware fake news detection with graph neural networks}.
\newblock \bibinfo{journal}{\emph{IEEE Transactions on Knowledge and Data Engineering}} (\bibinfo{year}{2023}).
\newblock


\bibitem[Xu and Kechadi(2023)]%
        {xu2023fuzzy}
\bibfield{author}{\bibinfo{person}{Cheng Xu} {and} \bibinfo{person}{M-Tahar Kechadi}.} \bibinfo{year}{2023}\natexlab{}.
\newblock \showarticletitle{Fuzzy Deep Hybrid Network for Fake News Detection}. In \bibinfo{booktitle}{\emph{Proceedings of the 12th International Symposium on Information and Communication Technology}}. \bibinfo{pages}{118--125}.
\newblock


\bibitem[Yang et~al\mbox{.}(2022)]%
        {yang2022coarse}
\bibfield{author}{\bibinfo{person}{Zhiwei Yang}, \bibinfo{person}{Jing Ma}, \bibinfo{person}{Hechang Chen}, \bibinfo{person}{Hongzhan Lin}, \bibinfo{person}{Ziyang Luo}, {and} \bibinfo{person}{Yi Chang}.} \bibinfo{year}{2022}\natexlab{}.
\newblock \showarticletitle{A Coarse-to-fine Cascaded Evidence-Distillation Neural Network for Explainable Fake News Detection}. In \bibinfo{booktitle}{\emph{Proceedings of the 29th International Conference on Computational Linguistics}}. \bibinfo{pages}{2608--2621}.
\newblock


\bibitem[Zhang et~al\mbox{.}(2020)]%
        {Zhang2020AnswerFactFC}
\bibfield{author}{\bibinfo{person}{Wenxuan Zhang}, \bibinfo{person}{Yang Deng}, \bibinfo{person}{Jing Ma}, {and} \bibinfo{person}{Wai Lam}.} \bibinfo{year}{2020}\natexlab{}.
\newblock \showarticletitle{{A}nswer{F}act: Fact Checking in Product Question Answering}. In \bibinfo{booktitle}{\emph{Proceedings of the 2020 Conference on Empirical Methods in Natural Language Processing (EMNLP)}}. \bibinfo{publisher}{Association for Computational Linguistics}, \bibinfo{address}{Online}, \bibinfo{pages}{2407--2417}.
\newblock
\urldef\tempurl%
\url{https://doi.org/10.18653/v1/2020.emnlp-main.188}
\showDOI{\tempurl}


\bibitem[Zhong et~al\mbox{.}(2020)]%
        {Zhong2020ReasoningOS}
\bibfield{author}{\bibinfo{person}{Wanjun Zhong}, \bibinfo{person}{Jingjing Xu}, \bibinfo{person}{Duyu Tang}, \bibinfo{person}{Zenan Xu}, \bibinfo{person}{Nan Duan}, \bibinfo{person}{Ming Zhou}, \bibinfo{person}{Jiahai Wang}, {and} \bibinfo{person}{Jian Yin}.} \bibinfo{year}{2020}\natexlab{}.
\newblock \showarticletitle{Reasoning Over Semantic-Level Graph for Fact Checking}. In \bibinfo{booktitle}{\emph{Proceedings of the 58th Annual Meeting of the Association for Computational Linguistics}}. \bibinfo{publisher}{Association for Computational Linguistics}, \bibinfo{address}{Online}, \bibinfo{pages}{6170--6180}.
\newblock
\urldef\tempurl%
\url{https://doi.org/10.18653/v1/2020.acl-main.549}
\showDOI{\tempurl}


\bibitem[Zhou et~al\mbox{.}(2019)]%
        {Zhou2019GEARGE}
\bibfield{author}{\bibinfo{person}{Jie Zhou}, \bibinfo{person}{Xu Han}, \bibinfo{person}{Cheng Yang}, \bibinfo{person}{Zhiyuan Liu}, \bibinfo{person}{Lifeng Wang}, \bibinfo{person}{Changcheng Li}, {and} \bibinfo{person}{Maosong Sun}.} \bibinfo{year}{2019}\natexlab{}.
\newblock \showarticletitle{{GEAR}: Graph-based Evidence Aggregating and Reasoning for Fact Verification}. In \bibinfo{booktitle}{\emph{Proceedings of the 57th Annual Meeting of the Association for Computational Linguistics}}. \bibinfo{publisher}{Association for Computational Linguistics}, \bibinfo{address}{Florence, Italy}, \bibinfo{pages}{892--901}.
\newblock
\urldef\tempurl%
\url{https://doi.org/10.18653/v1/P19-1085}
\showDOI{\tempurl}


\end{thebibliography}
